\pdfoutput=1

\documentclass[11pt]{article}

\usepackage[final]{acl}
\usepackage{times}
\usepackage{latexsym}
\usepackage[T1]{fontenc}
\usepackage[utf8]{inputenc}
\usepackage{microtype}
\usepackage{inconsolata}
\usepackage{graphicx}
\usepackage{amssymb}
\usepackage{algorithm}
\usepackage{algorithmicx}
\usepackage{algpseudocode}
\usepackage{amsmath}
\usepackage{booktabs}
\usepackage{graphicx}     
\usepackage{array}        
\usepackage{multirow}     
\usepackage{caption}      
\usepackage{enumitem}
\usepackage[normalem]{ulem}

\title{Subjective Behaviors and Preferences in \texttt{LLM}: Language of Browsing}

\author{
\normalfont
  Sai Sundaresan$^{1}$ \quad
  Harshita Chopra$^{2}$\footnotemark[2] \quad
  Atanu R. Sinha$^{1}$ \\
  Koustava Goswami$^{1}$ \quad
  Nagasai Saketh Naidu$^{1}$ \quad
  Raghav Karan$^{1}$\footnotemark[2] \quad
  N Anushka$^{1}$\footnotemark[2] \\
  $^{1}$Adobe Research \\
  $^{2}$University of Washington, Seattle \\
  \texttt{atr@adobe.com}
}

\begin{document}
\maketitle
\begin{abstract}
A Large Language Model (\texttt{LLM}) offers versatility across domains and tasks, purportedly benefiting users with a wide variety of behaviors and preferences. We question this perception about an \texttt{LLM} when users have inherently subjective behaviors and preferences, as seen in their ubiquitous and idiosyncratic browsing of websites or apps. The sequential behavior logs of pages, thus generated, form something akin to each user's self-constructed "language", albeit without the structure and grammar imbued in natural languages. We ask: (i) Can a small \texttt{LM} represent the "language of browsing" better than a large \texttt{LM}? (ii) Can an \texttt{LM} with a single set of parameters (or, single \texttt{LM}) adequately capture myriad users' heterogeneous, subjective behaviors and preferences? (iii) Can a single \texttt{LM} with high average performance, yield low variance in performance to make alignment good at user level? We introduce clusterwise \texttt{LM} training, \texttt{HeTLM} (Heterogeneity aware Training of Language Model), appropriate for subjective behaviors. We find that (i) a small \texttt{LM} trained using a page-level tokenizer outperforms large pretrained or finetuned \texttt{LMs}; (ii) \texttt{HeTLM} with heterogeneous cluster specific set of parameters outperforms a single \texttt{LM} of the same family, controlling for the number of parameters; and (iii) a higher mean and a lower variance in generation ensues, implying improved alignment.

\end{abstract}

\begin{figure}[t!]
\centering
\includegraphics[width=1\linewidth]{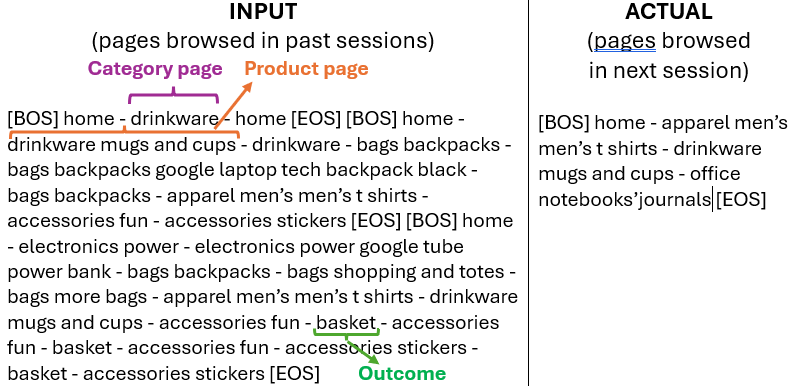}
\caption{\textit{Language of Browsing}: A user's 4 successive sessions are shown; 3 form the Input, 4th is Actual, used to compare  generation. ([BOS], [EOS]) demarcate sessions. Sequential page browsing is seen; sequence-length varies by sessions, and by users. Session's structure is a combination of \textit{Category Page, Product Page}, and \textit{Outcome} if it occurs. Outcome: Cart (Basket) or Purchase, is a firm desired target label, and occurs only in a few sessions.}
\label{fig:google-input-actual-eg}
\end{figure}

\section{Introduction}
\renewcommand{\thefootnote}{\fnsymbol{footnote}}
\footnotetext[2]{Work done while authors were at Adobe Research}
Large Language Models (\texttt{LLMs}) fuel expectations that a \textit{single} trained model can effectively align with preferences of myriad users for a given task within a domain. We term a \textit{single \texttt{LM}} as a specific \texttt{LM} with a \textit{single set} of parameters, regardless of its size. Examples of tasks in the domain of natural language are question-answer, summarization, etc.; and in the business domain - prediction of business metrics such as conversion, etc. Domain-mapping, task-calibration and preference alignment, can be presumably conquered through provisioning of specific context, instructions, examples, in-context learning, prompting, learning from human preference, finetuning or pretraining on domain-specific data, retrieval augmented generation, applied to a single \texttt{LM}. We question the paradigm of a single \texttt{LM} satisfying alignment for subjective behaviors of myriad users for a given task and domain.

Users' everyday interactions with websites and apps occur in an \textit{idiosyncratic} manner, generating a language of sequence of webpages (or, pages), termed \textit{language of browsing}. Fig~\ref{fig:google-input-actual-eg} shows an example of this. This language \textit{exemplifies} subjective behaviors and preferences, which are \textit{heterogeneous} and \textit{crucially} do not conform to oracle like behavior nor oracle-preference. Even if a single \texttt{LM} produces good average performance in generation across users, the variance in performance across users may be weak, making user level alignment weak. Thus, alignment with some \textit{ideal} oracle or with the average across users, is less meaningful. The language of browsing is ubiquitous in the form of pageurl-logs that online firms collect. We study users' heterogeneity of behaviors to examine the effectiveness of a single \texttt{LM}. Domain adaptation of \texttt{LM} addresses aspects of heterogeneity in domains and tasks, but none of these works tackle the important problem of heterogeneity in users' subjective behaviors and preferences, embodied in the sequential pages browsed. We learn the language intrinsic to such sequences, using decoder \texttt{LM}. For online firms, prediction of outcomes such as user conversion is of utmost importance. We thus consider three goals: (a) average performance in generation of pages, (b) variance of performance in generation of pages across users, and (c) performance on outcome prediction, where (b) and (c) are unattended in \texttt{LM} prior art. 

We find that small \texttt{LMs}, pretrained using a page-level tokenizer, outperform \texttt{GPT4o} with prompting, and fine-tuned versions of \texttt{LLaMa-3-8B, Mistral-7B, Gemma-7B}. The small \texttt{LMs}, are then finetuned clusterwise through our proposed \textbf{\texttt{HeTLM}} (\textbf{Heterogeneity aware Training of Language Model}), where both the number of clusters $K$ and assignment of users to clusters are \textit{endogenously} determined as finetuning progresses. The average performance in generation across users for \texttt{HeTLM} is better than the single \texttt{LM} of larger size in the same family, and the variance in performance is lower. Also, \texttt{HeTLM} performs well on prediction of outcomes available in the data--\textit{add to cart} and \textit{purchase}. These show that \texttt{HeTLM} improves performance for both page generation and outcome prediction. 

Our \textbf{contributions} are: \textbf{(i)} Drawing attention to Language of Browsing for \texttt{LLM}s; \textbf{(ii)} Showing that a small model can outperform a large model for outcome prediction with effective pretraining; \textbf{(iii)} Introducing heterogeneity of users' subjective behaviors and preferences in \texttt{LLM}s; \textbf{(iv)} Offering a network architecture to train \texttt{LLM} clusterwise. 

\section{Related Work}
\textbf{Heterogeneity:} Heterogeneity of subjective preferences inspires our inquiry~\cite{lu2022uncovering}. Heterogeneity drives research in clustering. Clustering algorithms have evolved with online data~\cite{aljalbout2018clustering,chen2017purtreeclust,ezenkwu2015application,alkhayrat2020comparative} - unsupervised and supervised~\cite{shin2019predictive,ghasemi2021survey}. However, our goal is not to cluster purely based on the user browsing sequence. \cite{lee2020temporalphenotypingusingdeep} learn discrete representations of time-series data by minimizing the KL divergence between individual and cluster-level outcome distributions. We extend this method to an LLM-based setting, where we train multiple cluster-specific \texttt{LLMs} so that at inference time, any arbitrary user's input can be given more accurate generation by invoking the \texttt{LLM} of the nearest cluster. To our best knowledge, such endogenous clusterwise \texttt{LLM} training and its use is not present in prior art. 

\noindent\textbf{Recommendation:} A growing body of work apply \texttt{LLM}s to recommendation tasks~\cite{wu2024survey,yang2023palr}. Their focus is on improving recommendations, but not on training \texttt{LLM}s for heterogeneous users with subjective preferences.  

\noindent\textbf{Domain Adaptation:} 
 Extensive research has addressed domain adaptation by pretraining \texttt{LLMs} on large, domain-specific datasets \citep[i.a.,][]{DBLP:journals/corr/abs-1904-03323,DBLP:conf/emnlp/LuDN22,lee2020biobert}, \texttt{BioBERT} \cite{DBLP:journals/bioinformatics/LeeYKKKSK20}, \texttt{BlueBERT} \cite{DBLP:conf/bionlp/PengYL19} , \texttt{BioClinicalBERT} \cite{DBLP:journals/corr/abs-1904-03323}, \texttt{diseaseBERT} \cite{DBLP:conf/emnlp/HeZZCC20}, \texttt{SciFive} \cite{DBLP:journals/corr/abs-2106-03598} and \texttt{BioBART} \cite{DBLP:conf/bionlp/YuanYGZXY22}. Subsequent work has explored fine-tuning these pretrained models on downstream tasks \cite{DBLP:journals/corr/abs-2306-09968,mishra2024llamat}.  
 Chen et al., \cite{DBLP:conf/emnlp/ChenHCCLY20} pointed out the complexity of adopting general language models blindly to downstream tasks by fine-tuning. Another line of research employs prompt learning for domain adaptation \citep{DBLP:conf/eacl/GoswamiLAA23}.
More recently, methods that infuse task-relevant information from related sources, rather than adapting to an entire domain, have shown promise \cite{DBLP:conf/icml/BorgeaudMHCRM0L22,DBLP:conf/iclr/DaiZMLNLBGHC23,DBLP:journals/corr/abs-2301-00303,DBLP:journals/corr/abs-2208-03299,DBLP:conf/nips/LewisPPPKGKLYR020}. He et al. \cite{DBLP:journals/corr/abs-2301-00303} proposed \texttt{LLM} based reasoning by decomposing tasks into multiple reasoning steps.

\section{Datasets} 
We show implementation on two public datasets\footnote{The processed datasets are provided here (\href{https://github.com/saisundaresan01/pageurl_prediction_emnlp}{link})}, each containing pageurls of consumer interactions. The pageurls are sequence of page-names browsed by each consumer, in each session, and sequenced by timestamp. To conserve space, all results for and the description of  Dataset II are available in Appendix~\ref{subsec:exp_data2}. Now, we confine to Dataset I. 

\textbf{Dataset I} \cite{google_analytics_sample}. Train : Test split = 47,274 : 5,253 samples. Number of unique pages = 1,123. We filter out very short sessions, incomplete data and extremely long sessions based on a percentile cutoff to create the dataset for experiments.

Fig.~\ref{fig:google-input-actual-eg} shows an example of Input data from Dataset I, as a sequence of pages. \textbf{Category Level}, \textbf{Product Level}, and \textbf{Outcome} pages form the tokens we use. \textbf{Outcomes} available in pageurls are: \textit{Cart} (Basket, in Dataset I) and \textit{Purchase}; these Outcome pages are generated by the \texttt{LM} and also used as \textit{target labels} to evaluate Outcome prediction against ground truth of the Actual (next) session. 

\section{Model}

\subsection{Small \texttt{LM} and Tokenization}\label{sec:choice-of-LM}
To study our thesis, we want an open, small \texttt{LM} which has larger sized variants available, as an open \texttt{LM} allows pretraining. To create a just comparison of the small variant, which when trained in a clusterwise manner has its total number of parameters increase by the multiple of the number of clusters, we need a large variant having at least the same number of total parameters. Since architectures differ by families of \texttt{LM}, it makes sense to select small and larger sized \texttt{LMs} of the same decoder family for a careful comparison. Also, the growing importance of the inference phase \cite{DBLP:journals/corr/abs-2408-03314} calls for a small \texttt{LM} with relatively few number of finalized clusters, so that the total number of parameters needed for inference does not blow up in \texttt{HeTLM}.    
Finally, a small, open \texttt{LM} allows for use of a custom tokenizer with a reduced vocabulary size, which can improve adherence to the desired output format. For this, we employ a custom tokenizer that performs page-level tokenization, including only the unique pages in the vocabulary. For \texttt{HeTLM} clusterwise training we use three small \texttt{LMs}: \texttt{OPT-350M}, \texttt{QWEN-2.5-500M}, and \texttt{SmolLM2-360M}. 

\subsection{Exogenous Clusterwise Training: \texttt{Kmeans}}
User sessions are clustered using their \texttt{SBERT} embeddings, with fixed $K$, using the \texttt{Kmeans} algorithm. Then, cluster-wise, $K$ different small \texttt{LMs} of the same family are finetuned. This approach, which also recognizes the heterogeneity of users, is straightforward and a useful baseline. 
The limitations are that $K$ is fixed and user sessions are assigned to clusters based purely on embeddings, and are not re-assigned endogenously as proposed in \texttt{HeTLM}, which is described next.   

\subsection{Endogenous Clusterwise Training: \texttt{HeTLM}}
\begin{figure}[t]
    \centering
    \includegraphics[width=0.95\columnwidth]{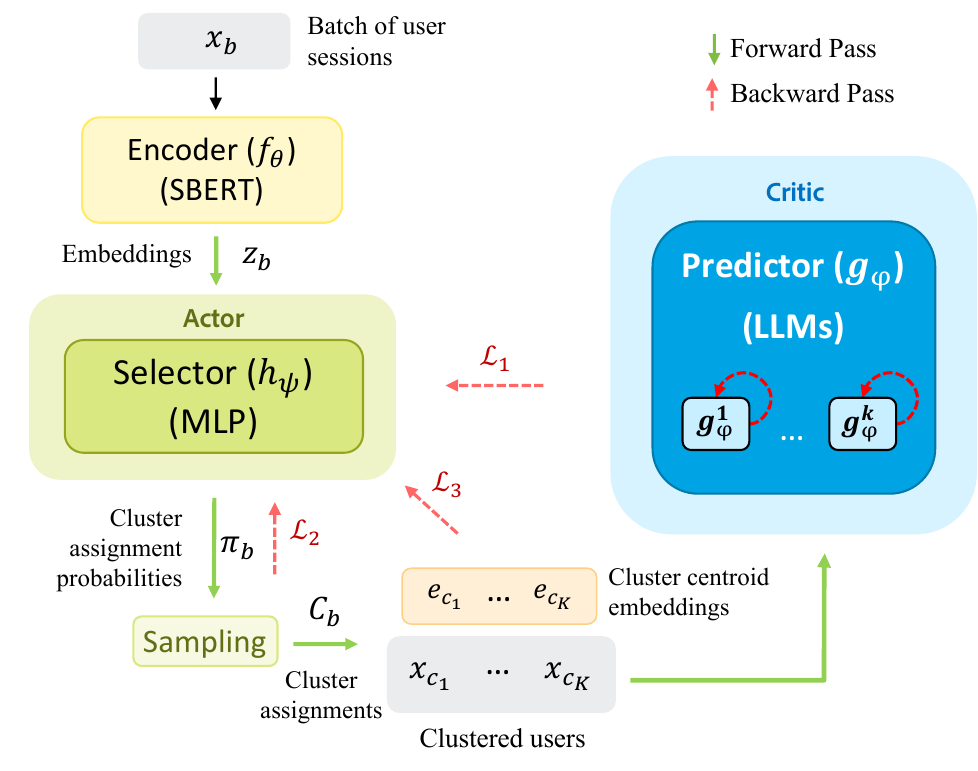}
    \caption{\texttt{HeTLM} Architecture}
    \label{fig:arch_diag}
\end{figure}

We propose \textbf{\texttt{HeTLM}} (\textbf{He}terogeneity aware \textbf{T}raining of \textbf{L}anguage \textbf{M}odels). Fig.~\ref{fig:arch_diag} shows the architecture of \textbf{\texttt{HeTLM}} which integrates embedding-based clustering and fine-tuning endogenously, where clustering is informed by fine-tuning and fine-tuning is guided by clustering. By clustering user session embeddings and fine-tuning a dedicated \texttt{LM} for each cluster, it captures user-specific patterns more effectively than a single-model or exogenous methods like K-means. We use an Actor–Critic framework to iteratively refine both the clustering and the number of clusters, $K$, based on prediction quality. The theoretical basis for our method is given in~\ref{sec:theoretical-basis}. The model has three components: the \textit{Encoder} (\texttt{SBERT} \cite{reimers2019sentencebertsentenceembeddingsusing}), the \textit{Selector} (\texttt{MLP}), and the \textit{Predictor}~\cite{zhang2022optopenpretrainedtransformer}). A \texttt{small-LM} serves as the \textit{Predictor}. 

\noindent(1) The \textit{Encoder} processes a batch of user sessions ($x_b$) and generates embedding representations ($z_b$) for each sample. These embeddings serve as input to the \textit{Selector}. We choose \texttt{SBERT} for the encoder due to its efficiency in generating high-quality embeddings for sentence-level data.

\noindent (2) The \textit{Selector} takes these embeddings and generates a probability distribution over $K$ clusters for each user session. It is pretrained to mimic \texttt{Kmeans} cluster assignments. Based on these probabilities, the user sessions are grouped and passed to the corresponding \textit{Predictor} models. The Selector, an \texttt{MLP}, efficiently maps session embeddings to clusters, where the number of clusters is \textit{endogenous}. 

\noindent (3) The \textit{Predictor} consists of $K$ instances of the small \texttt{LM} model, each pretrained on the entire dataset. We use a custom tokenizer with a vocabulary constrained to the set of pages in the dataset, resulting in better adherence to the output format. Each Predictor is then fine-tuned only on the user sessions assigned to its corresponding cluster by the Selector. By allowing a different set of parameter weights for each cluster, we allow each cluster-wise \texttt{LM} to specialize in behavior patterns of its assigned users, improving predictive accuracy. 

\noindent (4) The \textit{Selector} is the \textit{Actor}, and the \textit{Predictor} is the critic. The Selector's cluster assignments are refined through three loss functions, defined in Sec~\ref{sec:llm-loss} and~\ref{sec:all-losses}. 
Loss $\mathcal{L}_1$, is the negative log-likelihood (NLL) of each Predictor on its assigned sessions, which encourages assignments that yield strong specialization. Loss $\mathcal{L}_2$, is the NLL of the cluster assignment probabilities which encourages a sharper probability distribution. 
Loss, $\mathcal{L}_3$, is obtained from the Manhattan distance between centroid embeddings of pairwise clusters, promoting separation between the clusters. Description of the losses and training procedure follow next.

\subsection{\texttt{LLM} Loss}\label{sec:llm-loss}

Each sequence of user sessions consists of pages separated by delimiters. Given a target sequence of pages \( x = a_1 - a_2 - \dots - a_m \), we define its token representation as \( T = t_1, t_2, \dots, t_n \), where \( n \) is the number of tokens. The corresponding one-hot encoded matrix is given by \( \mathfrak{O} = [O_1, O_2, \dots, O_n] \), where \( O_i \) is one-hot vector of length equal to the vocabulary size, representing the token $t_i$, and \( n \) is the token sequence length.

A language model generates probability distributions corresponding to each token, represented as \( \mathfrak{P} = [P_1, P_2, \dots, P_n] \), where each \( P_i \) is a probability distribution for the next token over the vocabulary. The loss function is computed as the sum of negative log-likelihoods (\texttt{NLL}) between the one-hot encoded \( O_i \) and the generated probability distribution \( G_i \): $\mathcal{L}_{LLM} = - \sum_{i=1}^{n} O_i \log P_i$

\subsection{\texttt{HeTLM}: Actor-Critic}\label{sec:all-losses}

Our proposed \texttt{HeTLM} method follows an Actor-Critic paradigm, where an Encoder first extracts user representations, and then an Actor (Selector) and Critic (\texttt{LLM}) operate on these embeddings. Let $\mathcal{X}$ represent the user sessions in the dataset and $\mathcal{Z}$ represent the embedding space of the Encoder. Let $V$ be the vocabulary size of the model.

\textbf{Encoder} \(f_\theta:\mathcal{X} \to \mathcal{Z} \) maps a user’s session sequence \( x \in \mathcal{X} \) to a session embedding \( z \in \mathcal{Z} \). We utilize the \texttt{SBERT} model to produce these embeddings. The hidden vector \( z \) represents the latent tendency of a user, and is used for clustering. The Encoder weights are kept frozen.

\textbf{Selector} \( h_\psi: \mathcal{Z} \to \Delta^{K-1} \) is a \texttt{MLP} that assigns \( z \) to one of \( K \) clusters, computing a probability distribution \( \pi \) where \( \pi(k) \) is the probability of assigning \( z \) to cluster \( k \).  

\textbf{Predictor} \( g_\phi^k: \mathcal{X} \to \mathbb{R}^{n \times V} \) is an \texttt{LLM} which maps a user session corresponding to cluster k to a matrix, where each row represents a probability distribution $P_i \in \Delta^{V-1}$ for each token.

\textbf{Embedding Dictionary} \( \mathcal{E} \): stores the centroids of the session embeddings of \( K \) clusters. Given a cluster assignment \( c_k \), it stores the corresponding centroid embedding \( e_{c_k} \in \mathcal{H} \).

\subsection{Losses and Training} \label{Sec:Training}
The following steps are used:

\begin{enumerate}[nosep]
        \item Generate Encoder embeddings $z$ for all $x \in \mathcal{X}$
	\item Initialize the cluster embeddings using \texttt{K-means} on the embeddings $z$ and pre-train the Selector to mimic \texttt{K-means}
        \item Pretrain each of the Predictors on all $x \in \mathcal{X}$
        \item Perform Actor-Critic Finetuning to iteratively improve the Selector and the Predictors
\end{enumerate}

\begin{figure}[t]
    \centering
    \includegraphics[width=.8\columnwidth]{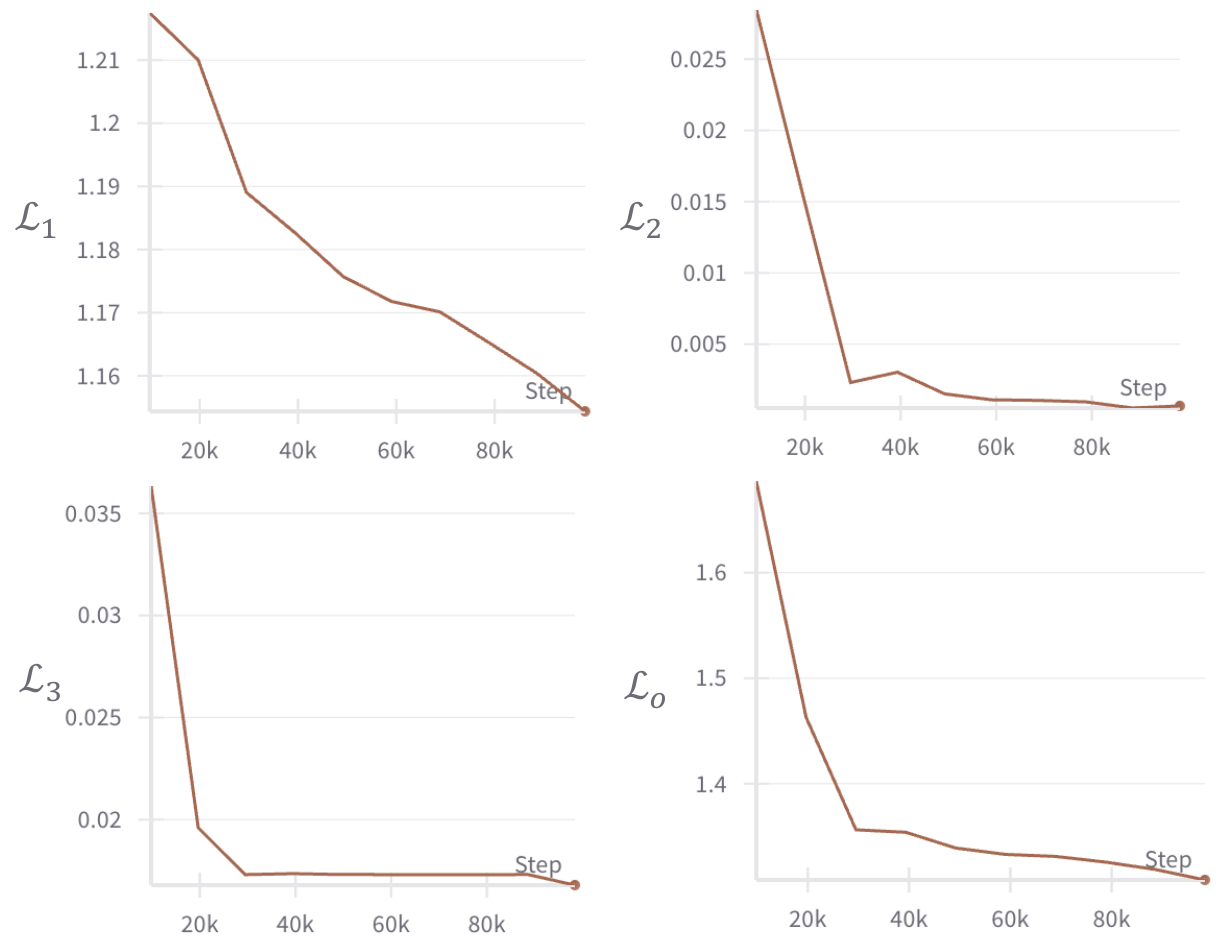}
    \caption{Dataset I. Validation losses for $\alpha=5,\beta=9$}
    \label{fig:val-losses}
\end{figure}

The overall selector loss $\mathcal{L}_O (\theta, \phi, \psi) = \mathcal{L}_1 + \alpha \mathcal{L}_2 + \beta \mathcal{L}_3$ combines the 3 losses as described below, with $\alpha$ and $\beta$ as hyperparameters.

The loss term $\mathcal{L}_1$ ensures effective specialization within the predictors. We take the weighted average over the cluster probabilities of the \texttt{LLM} loss for all $x \in \mathcal{X}$ to allow for backpropagation.

\begin{equation}
 \mathcal{L}_1 (\theta, \phi, \psi) = \sum_{k \in K} \pi_k \big[l_1^k \big]
 \label{critic_loss}
\end{equation}

where, ${l_1^k}$ = $\mathcal L_{LLM}$, is the negative log likelihood of the \texttt{LLM} $g_{\phi}^k$. 
The critic's loss, which is the loss for each of the Predictors, is:

\begin{equation}
 \mathcal{L}_c^k (\phi) = \mathbf{1}_{\{argmax(\pi) = k\}} \big[l_1^k \big]
 \label{critic_loss}
\end{equation}

The loss term $\mathcal{L}_2$ promotes sparse cluster assignment such that each user belongs to one cluster with high probability.
It is given by:
\begin{equation}
	\mathcal{L}_2 (\theta, \psi) = -\sum_{k \in K}\pi_k \log \pi_k
	\label{pis_cross_entropy_loss}
\end{equation}

Further, to promote well-separated cluster centroids in embedding representation, the loss $\mathcal{L}_3$ is used, which is the negated sum of Manhattan distance between pairwise cluster centroid embeddings passed through a sigmoid function,

\begin{equation}
    \mathcal{L}_3 (\theta, \phi, \psi) = \sigma\bigg(-\sum_{k \neq k'} \|\mathbf{e_{c_k}} - \mathbf{e_{c_{k'}}}\|_1 \bigg)
    \label{embedding_loss}
\end{equation}

For efficiency, training is performed in batches. The validation losses over steps are shown in Fig.~\ref{fig:val-losses} and the training steps are shown in Algorithm \ref{alg:actor_critic}. Training hyperparameters are included in~\ref{subsec:exp_setup}.

\begin{algorithm}
\small
\caption{Actor-Critic Fine-Tuning}
\label{alg:actor_critic}
\begin{algorithmic}[1]
\State \textbf{Generate:} embeddings z = $f_\theta(x)$ for all $x \in \mathcal{X}$
\State \textbf{Initialize:} Selector $h_{\psi}$ with \texttt{K-means} clusters
\State \textbf{Pretrain:} Predictors $g_\phi^k$ on all $x \in \mathcal{X}$
\State Train $h_\psi$, $g_{\phi}^k$ using Actor-Critic Finetuning
\While{not converged}
    \State sample batch $x_b \in \mathcal{X}$
    \State Compute embeddings $\mathbf{z_b} = f_\theta(\mathbf{x_b})$
    \State Compute cluster probs $\pi_b = h_\psi(\mathbf{z_b})$
    \State Assign clusters $C_b = argmax(\pi_b)$
    \State Critic loss: $\mathcal{L}_C = \mathcal{L}_1$
    \State Update $\phi$ via gradient descent on $\mathcal{L}_C$
    \State Selector loss: $\mathcal{L}_O = \mathcal{L}_1 + \alpha \mathcal{L}_2 + \beta \mathcal{L}_3$
    \State Update $\psi$ via gradient descent on $\mathcal{L}_O$
\EndWhile
\end{algorithmic}
\end{algorithm} 
\section{Empirical Strategy for Experiments}\label{sec:empirical-strategy}
We adopt a 2-step empirical strategy. 

\noindent\textbf{Step 1: Compare single \texttt{LMs} - small vs. large.}\label{emp-step-1} On Dataset I, as appropriate, each \texttt{LM} is--pretrained, or finetuned, or prompt-tuned--as the case may be, and compared on Mean performance metrics conforming to both Page Generation and Outcome Prediction for the Next session, across all users in test data. The performance of the small \texttt{LMs} is checked.

\noindent\textbf{Step 2: Compare \texttt{HeTLM} clusterwise, small \texttt{LM} vs. larger, single \texttt{LM}.}\label{emp-step-2}  Using Dataset I, each small \texttt{LM} is pretrained with the \texttt{HeTLM} architecture (Fig.~\ref{fig:arch_diag}) to produce $K$ clusterwise versions of this \texttt{LM}, where both $K$ and assignment of users to clusters are endogenously determined. We get $K$ \textit{different} sets of parameters for $K$ \texttt{LMs}, meant to better align overall with users in $K$ clusters. Fetch an \texttt{LM} of the same family as the small \texttt{LM}, but one with a larger size, so that its number of parameters is larger than the total number of parameters of $K$ small \texttt{LMs}. Pre-train the large \texttt{LM} with a page-level custom tokenizer on Dataset I. Compare performance of \texttt{HeTLM} clusterwise, small \texttt{LMs} with the larger single \texttt{LM} in average performance metrics, as well as, in Variance of performance in Page Generation across all users in test data. For indepth comparison, we use an expanded set of metrics.  

\subsection{Evaluation Metrics for \textit{Step 1} } \label{eval-metrics-detail}
All evaluations are performed by comparing page generation in next session against the pages in the Actual next session (e.g., Fig.~\ref{fig:google-input-actual-eg}). We use a broad set of metrics to support three key aspects of our approach. (1) Variance metrics in page generation are introduced to capture variability across users—an aspect overlooked in prior work but central to our premise of subjectivity in user heterogeneity. This aligns with our thesis of clusterwise \texttt{LMs} and is described in~\ref{variance-metric}. (2) Page generation metrics are aligned with prior work on language models and allow standard benchmarking. (3) Outcome prediction metrics (Accuracy, Precision, Recall, F1) are widely used in prior art and are essential for evaluating business impact. We further disaggregate these by outcome type (e.g., cart vs. purchase) to give the reader a more complete view of model behavior. Rather than reducing to a minimal set, we provide comprehensive metrics so readers can interpret results according to their own priorities. To aid in comparison across models, we also include composite metrics in Sec. 5.3. In Step 1, our first level comparison, we use 6 evaluation metrics. In Step 2, we use 14 \textit{additional} metrics to deep dive for detailed comparisons. 

\subsubsection{Page Generation - Mean across Users} 
A common objective for \texttt{LLMs}, we use the following 2 metrics in Step 1, which are defined per user. We report the Mean across all users, \textit{higher is better}.

\noindent\textbf{[IoA]} \textit{Intersection over Actual}: Ratio of number of correctly generated pages to that of \textit{actual} pages; same as, $1$ \textit{minus} \textit{False Negative}. 

\noindent\textbf{[IoP]} \textit{Intersection over Predicted (Generated)}: Ratio of number of correctly generated pages to that of \textit{generated} pages; same as, $1$ \textit{minus} \textit{False Positive}. 

\subsubsection{Outcome Prediction - Mean across Users} 
Important business Outcomes of interest are: Add-to-cart and Purchase. For each consumer we check whether generated pages in the Next session contain either Cart or Purchase page, and evaluate against the Actual session pages containing them. We use 4 outcome metrics, defined as Mean across all users. While F1 is a composite, recall and precision provide useful, incremental information and thus we use them. \textit{Higher is better}.

\noindent\textbf{[Acc]} \textit{Accuracy for Cart / Purchase}. 

\noindent\textbf{[Rec]} \textit{Recall for Cart / Purchase}. 

\noindent\textbf{[Prec]} \textit{Precision for Cart / Purchase}. 

\noindent\textbf{[F1]}   \textit{F1-score for Cart / Purchase}. 

\subsection{\textit{Additional} Evaluation Metrics for Step 2} \label{addl-metrics-step-2-3}

\subsubsection{Page Generation - Mean across Users}
We use the following 4 \textit{additional} metrics in Step 2. Each metric is computed per user and then reported as Mean across all users. \textit{Higher is better}. 

\noindent\textbf{[HR]} \textit{Hit Rate}: Per user, at least one common page in generation and actual is scored 1; 0, otherwise. 

\noindent\textbf{[IoU]} \textit{Intersection over Union}: Per user, ratio of the number of correctly generated pages to the union of the number of \textit{generated} and \textit{actual} pages. 

\noindent\textbf{[Val-P]} \textit{Valid Page Score}: Per user, ratio of the number of valid pages to the number of all pages in generation, per user. This cautions about \textit{hallucination}, by having a score closer to 1. 

\noindent\textbf{[New-P]} \textit{New Page Score}: Per user, ratio of the number of pages not in input to the number of valid pages in generation. Indicating a degree to which pages generated are outside of the pages in input, it captures new generation instead of merely retrieval. 

\subsubsection{\textbf{Introducing Variance in Page Generation metrics for User alignment}}\label{variance-metric} Per user, we can measure alignment since the Ground truth of Next session is available in the Language of Browsing. For the following metrics we compute Variance across users as a degree of alignment in Page Generation. \textit{Lower variance is better}, since it suggests better alignment.  

\noindent\textbf{[IoA-var]} \textit{Intersection over Actual Variance}.
        
\noindent\textbf{[IoG-var]} \textit{Intersection over Generated Variance}. 
        
\noindent\textbf{[IoU-var]} \textit{Intersection over Union Variance}.

\noindent\textbf{[HR-var]} \textit{Hit Rate Variance}.  

\subsubsection{Outcome Prediction separately for Cart and for Purchase} For a more complete picture to emerge, we also consider Accuracy, Recall, Precision separately for Cart and for Purchase predictions.   

\subsection{Composite Metrics}\label{sec:composie-metrics} To make comparison easier over the 20 metrics, we introduce composite metrics, where each is a scalar value . Defined with competing Single \texttt{LM} as benchmark, it is = \textit{{Number of metrics where \texttt{HeTLM} scores greater than single \texttt{LM}} / $20$}. This metric is intuitive, bounded between 0 and 1, where score greater (\textit{less}) than 0.5 favors (\textit{disfavors}) \texttt{HeTLM}, versus single \texttt{LLM}. For each of Page Generation-Mean, Outcome Prediction-Mean, Page Generation-Variance, and for Overall (across all these three metrics) we present a composite metric.  

\section{Experiment Results: Dataset I}\label{results-goog-step1}

\begin{table}[!h]
\centering

\resizebox{\columnwidth}{!}{
\begin{tabular}{l|cccccc}
\toprule
Model & IoA & IoP & Acc & Rec & Prec & F1 \\
    {} & {} & {} & Cart or & Cart or & Cart or & Cart or \\
    {} & {} & {} & Purchase & Purchase & Purchase & Purchase \\
\midrule
\texttt{GPT-4o Zero-shot} & 0.275 & 0.175 & 0.928 & 0.400 & 0.488 & 0.440 \\ \midrule
\texttt{GPT-4o Few-shot} & 0.400 & 0.276 & 0.869 & \textbf{0.628} & 0.367 & 0.464 \\ \midrule
\texttt{Llama2-7B Chat-Few-shot} & 0.329 & 0.251 & \textbf{0.932} & 0.305 & 0.424 & 0.355 \\ \midrule
\texttt{Llama3-8B LORA-32R} & 0.327 & 0.339 & 0.914 & 0.390 & 0.409 & 0.399 \\ \midrule
\texttt{Llama3-8B FullFineTune} & 0.295 & 0.307 & 0.871 & 0.399 & 0.437 & 0.417 \\ \midrule
\texttt{Mistral-7B LORA-32R} & 0.124 & 0.129 & 0.904 & 0.360 & 0.376 & 0.368 \\ \midrule
\texttt{Gemma-7B LORA-32R} & 0.269 & 0.266 & 0.905 & 0.442 & 0.424 & 0.433 \\ \midrule
\texttt{OPT-350M Pre-train} & 0.334 & 0.314 & 0.872 & 0.447 & 0.514 & 0.478 \\  \midrule
\texttt{QWEN-2.5-500M Pre-train} & 0.408 & \textbf{0.426} & 0.918 & 0.515 & \textbf{0.571} & \textbf{0.542} \\  \midrule
\texttt{SmolLM2-360M Pre-train} & \textbf{0.431} & \textbf{0.424} & \textbf{0.92} & 0.502 & \textbf{0.542} & \textbf{0.521} \\

\bottomrule
\end{tabular}
}
\caption{
\textbf{Dataset I}. Small 
\texttt{LM}s, \texttt{QWEN-2.5-500M}, \texttt{SmolLM2-360M}, \texttt{OPT-350M} dominate the collection of large \texttt{LMs} in 5 metrics, except in Rec - Cart or Purchase.}
\label{tab:results-baseline-chatgpt}
\end{table}

\begin{table}[bth!]
  
\resizebox{\columnwidth}{!}{ 
\begin{tabular}{@{}lccccccc@{}}
\toprule
\multicolumn{1}{c|}{\textbf{Model}} & \multicolumn{1}{c|}{\textbf{$N$}} & \textbf{\begin{tabular}[c]{@{}c@{}}HR \end{tabular}} & \textbf{\begin{tabular}[c]{@{}c@{}}IoA\end{tabular}} & \textbf{\begin{tabular}[c]{@{}c@{}}IoP\end{tabular}} & \textbf{\begin{tabular}[c]{@{}c@{}}IoU\end{tabular}} & \textbf{\begin{tabular}[c]{@{}c@{}}New-P\end{tabular}} & \textbf{\begin{tabular}[c]{@{}c@{}}Val-P\end{tabular}} \\ \midrule
\multicolumn{1}{l|}{\textbf{OPT 2.7B}} & \multicolumn{1}{c|}{5253} & 0.813 & 0.424 & 0.417 & 0.31 & 0.183 & 0.035 \\ \midrule
\multicolumn{8}{l}{\textbf{\texttt{OPT-350M Kmeans}, K=6}} \\ \midrule
\multicolumn{1}{l|}{Cluster 1} & \multicolumn{1}{c|}{438} & 0.838 & 0.419 & 0.31 & 0.228 & 0.196 & 0.028 \\
\multicolumn{1}{l|}{Cluster 2} & \multicolumn{1}{c|}{1359} & 0.703 & 0.221 & 0.165 & 0.104 & 0.018 & 0.003 \\
\multicolumn{1}{l|}{Cluster 3} & \multicolumn{1}{c|}{638} & 0.803 & 0.327 & 0.26 & 0.169 & 0.205 & 0.035 \\
\multicolumn{1}{l|}{Cluster 4} & \multicolumn{1}{c|}{330} & 0.745 & 0.472 & 0.451 & 0.362 & 0.364 & 0.078 \\
\multicolumn{1}{l|}{Cluster 5} & \multicolumn{1}{c|}{573} & 0.696 & 0.277 & 0.341 & 0.209 & 0.265 & 0.071 \\
\multicolumn{1}{l|}{Cluster 6} & \multicolumn{1}{c|}{1915} & 0.794 & 0.309 & 0.301 & 0.188 & 0.241 & 0.049 \\
\multicolumn{1}{l|}{Combined} & \multicolumn{1}{c|}{5253} & 0.761 & 0.304 & 0.275 & 0.18 & 0.186 & 0.038 \\ \midrule
\multicolumn{8}{l}{\textbf{\texttt{OPT-350M Kmeans}, K=2}} \\ \midrule
\multicolumn{1}{l|}{Combined} & \multicolumn{1}{c|}{5253} & 0.776 & 0.347 & 0.324 & 0.225 & 0.134 & 0.027 \\ \midrule
\multicolumn{8}{l}{\textbf{\texttt{OPT-350M HeTLM} ($\alpha$=2, $\beta$=1)}} \\ \midrule
\multicolumn{1}{l|}{Cluster 1} & \multicolumn{1}{c|}{948} & 0.795 & 0.355 & 0.354 & 0.24 & 0.172 & 0.033 \\
\multicolumn{1}{l|}{Cluster 2} & \multicolumn{1}{c|}{319} & 0.881 & 0.63 & 0.657 & 0.538 & 0.0 & 0.0 \\
\multicolumn{1}{l|}{Cluster 3} & \multicolumn{1}{c|}{2875} & 0.8 & 0.367 & 0.396 & 0.264 & 0.181 & 0.042 \\
\multicolumn{1}{l|}{Cluster 4} & \multicolumn{1}{c|}{277} & 0.686 & 0.298 & 0.406 & 0.245 & 0.051 & 0.016 \\
\multicolumn{1}{l|}{Cluster 5} & \multicolumn{1}{c|}{759} & 0.783 & 0.288 & 0.334 & 0.188 & 0.159 & 0.043 \\
\multicolumn{1}{l|}{Combined} & \multicolumn{1}{c|}{5253} & 0.796 & 0.367 & 0.397 & 0.265 & 0.156 & 0.036 \\ \midrule
\multicolumn{8}{l}{\textbf{\texttt{OPT-350M HeTLM} ($\alpha$=5, $\beta$=9)}} \\ \midrule
\multicolumn{1}{l|}{Cluster 1} & \multicolumn{1}{c|}{305} & 0.879 & 0.616 & 0.713 & 0.572 & 0.0 & 0.0 \\
\multicolumn{1}{l|}{Cluster 2} & \multicolumn{1}{c|}{4797} & 0.819 & 0.373 & 0.387 & 0.265 & 0.216 & 0.048 \\
\multicolumn{1}{l|}{Cluster 3} & \multicolumn{1}{c|}{151} & 0.57 & 0.31 & 0.439 & 0.286 & 0.0 & 0.0 \\
\multicolumn{1}{l|}{Combined} & \multicolumn{1}{c|}{5253} & \textbf{0.816} & \textbf{0.385} & \textbf{0.407} & \textbf{0.284} & \textbf{0.197} & \textbf{0.044} \\ 
\midrule
\midrule
\midrule
\multicolumn{1}{l|}{\textbf{QWEN-2.5-7B}} & \multicolumn{1}{c|}{5253} & 0.699 & 0.253 & 0.265 & 0.156 & 0.136 & 0.044 \\
\midrule
\multicolumn{8}{l}{\textbf{\texttt{QWEN-2.5-500M HeTLM} ($\alpha$=5, $\beta$=9)}} \\ \midrule
\multicolumn{1}{l|}{Cluster 1} & \multicolumn{1}{c|}{3101} & 0.809 & 0.437 & 0.39 & 0.30 & 0.223 & 0.041 \\
\multicolumn{1}{l|}{Cluster 2} & \multicolumn{1}{c|}{316} & 0.87 & 0.64 & 0.71 & 0.587 & 0.022 & 0.005 \\
\multicolumn{1}{l|}{Cluster 3} & \multicolumn{1}{c|}{1673} & 0.80 & 0.401 & 0.414 & 0.30 & 0.175 & 0.037 \\
\multicolumn{1}{l|}{Cluster 4} & \multicolumn{1}{c|}{163} & 0.595 & 0.322 & 0.391 & 0.27 & 0.006 & 0.001 \\
\multicolumn{1}{l|}{Combined} & \multicolumn{1}{c|}{5253} & 0.803 & 0.434 & 0.417 & 0.314 & 0.189 & 0.036 \\ 
\bottomrule
\end{tabular}
}
\caption{Dataset I. Page Generation Results. \textit{Higher is Better}. \textbf{Combined} shows the average across clusters for each \texttt{HeTLM}.}
\label{goog-data-step2-page-gen-compare}
\end{table}

\begin{table}[h!]
\resizebox{\columnwidth}{!}{%
\begin{tabular}{@{}lccccc@{}}
\toprule
\multicolumn{1}{c|}{\textbf{Model}} & \multicolumn{1}{c|}{\textbf{$N$}} & \textbf{\begin{tabular}[c]{@{}c@{}}HR-var\end{tabular}} & \textbf{\begin{tabular}[c]{@{}c@{}}IoA-var\end{tabular}} & \textbf{\begin{tabular}[c]{@{}c@{}}IoP-var\end{tabular}} & \textbf{\begin{tabular}[c]{@{}c@{}}IoU-var\end{tabular}} \\ \midrule
\multicolumn{1}{l|}{OPT 2.7B} & \multicolumn{1}{c|}{5253} & 0.152 & 0.121 & 0.119 & 0.109 \\ \midrule
\multicolumn{6}{l}{\textbf{OPT-350M: Kmeans, K=6}} \\ \midrule
\multicolumn{1}{l|}{Cluster 1} & \multicolumn{1}{c|}{438} & 0.136 & 0.088 & 0.076 & 0.053 \\
\multicolumn{1}{l|}{Cluster 2} & \multicolumn{1}{c|}{1359} & 0.209 & 0.047 & 0.031 & 0.015 \\
\multicolumn{1}{l|}{Cluster 3} & \multicolumn{1}{c|}{638} & 0.158 & 0.068 & 0.06 & 0.031 \\
\multicolumn{1}{l|}{Cluster 4} & \multicolumn{1}{c|}{330} & 0.19 & 0.155 & 0.148 & 0.129 \\
\multicolumn{1}{l|}{Cluster 5} & \multicolumn{1}{c|}{573} & 0.211 & 0.087 & 0.108 & 0.075 \\
\multicolumn{1}{l|}{Cluster 6} & \multicolumn{1}{c|}{1915} & 0.163 & 0.069 & 0.07 & 0.044 \\
\multicolumn{1}{l|}{Combined} & \multicolumn{1}{c|}{5253} & 0.182 & 0.077 & 0.074 & 0.048 \\ \midrule
\multicolumn{6}{l}{\textbf{OPT-350M: Kmeans, K=2}} \\ \midrule
\multicolumn{1}{l|}{Combined} & \multicolumn{1}{c|}{5253} & 0.174 & 0.096 & 0.092 & 0.071 \\ \midrule
\multicolumn{6}{l}{\textbf{\texttt{OPT-350M: HeTLM} ($\alpha$=2, $\beta$=1)}} \\ \midrule
\multicolumn{1}{l|}{Cluster 1} & \multicolumn{1}{c|}{948} & 0.163 & 0.094 & 0.09 & 0.071 \\
\multicolumn{1}{l|}{Cluster 2} & \multicolumn{1}{c|}{319} & 0.105 & 0.138 & 0.122 & 0.129 \\
\multicolumn{1}{l|}{Cluster 3} & \multicolumn{1}{c|}{2875} & 0.16 & 0.1 & 0.102 & 0.083 \\
\multicolumn{1}{l|}{Cluster 4} & \multicolumn{1}{c|}{277} & 0.215 & 0.102 & 0.133 & 0.091 \\
\multicolumn{1}{l|}{Cluster 5} & \multicolumn{1}{c|}{759} & 0.17 & 0.061 & 0.072 & 0.037 \\
\multicolumn{1}{l|}{Combined} & \multicolumn{1}{c|}{5253} & 0.162 & 0.101 & 0.103 & 0.082 \\ \midrule
\multicolumn{6}{l}{\textbf{\texttt{OPT-350M: HeTLM} ($\alpha$=5, $\beta$=9)}} \\ \midrule
\multicolumn{1}{l|}{Cluster 1} & \multicolumn{1}{c|}{305} & 0.107 & 0.137 & 0.129 & 0.144 \\
\multicolumn{1}{l|}{Cluster 2} & \multicolumn{1}{c|}{4797} & 0.148 & 0.096 & 0.097 & 0.082 \\
\multicolumn{1}{l|}{Cluster 3} & \multicolumn{1}{c|}{151} & 0.245 & 0.143 & 0.185 & 0.137 \\
\multicolumn{1}{l|}{Combined} & \multicolumn{1}{c|}{5253} & 0.15 & 0.103 & 0.107 & 0.092 \\ 
\midrule
\midrule
\midrule
\multicolumn{1}{l|}{QWEN-2.5-7B} & \multicolumn{1}{c|}{5253} & 0.211 & 0.064 & 0.067 & 0.036 \\ \midrule
\multicolumn{6}{l}{\textbf{QWEN-2.5-500M: \texttt{HeTLM} ($\alpha$=5, $\beta$=9)}} \\ \midrule
\multicolumn{1}{l|}{Cluster 1} & \multicolumn{1}{c|}{3101} & 0.154 & 0.125 & 0.103 & 0.098 \\
\multicolumn{1}{l|}{Cluster 2} & \multicolumn{1}{c|}{316} & 0.113 & 0.143 & 0.135 & 0.151 \\
\multicolumn{1}{l|}{Cluster 3} & \multicolumn{1}{c|}{1673} & 0.161 & 0.123 & 0.115 & 0.109 \\
\multicolumn{1}{l|}{Cluster 4} & \multicolumn{1}{c|}{163} & 0.241 & 0.14 & 0.170 & 0.128 \\
\multicolumn{1}{l|}{Combined} & \multicolumn{1}{c|}{5253} & 0.158 & 0.129 & 0.117 & 0.111 \\ 
\bottomrule
\end{tabular}%
}
\caption{Dataset I. Variance in Page Generation across users. \textit{Lower is Better}. Results for same models as in Table~\ref{goog-data-step2-outcome-pred-compare}.}
\label{goog-data-step2-page-gen-variance}
\end{table}

\begin{table*}[t!]
\resizebox{\textwidth}{!}{  
\begin{tabular}{@{}lccccccccccc@{}}
\toprule
\multicolumn{1}{c|}{\textbf{Model}} & \multicolumn{1}{c|}{\textbf{\begin{tabular}[c]{@{}c@{}}N\end{tabular}}} & \textbf{\begin{tabular}[c]{@{}c@{}}Acc-Cart\end{tabular}} & \textbf{\begin{tabular}[c]{@{}c@{}}Acc-Purchase\end{tabular}} & \textbf{\begin{tabular}[c]{@{}c@{}}Acc-Cart/Purchase\end{tabular}} & \textbf{\begin{tabular}[c]{@{}c@{}}Rec-Cart\end{tabular}} & \textbf{\begin{tabular}[c]{@{}c@{}}Rec-Purchase\end{tabular}} & \textbf{\begin{tabular}[c]{@{}c@{}}Rec-Cart/Purchase\end{tabular}} & \textbf{\begin{tabular}[c]{@{}c@{}}Prec-Cart\end{tabular}} & \textbf{\begin{tabular}[c]{@{}c@{}}Prec-Purchase\end{tabular}} & \textbf{\begin{tabular}[c]{@{}c@{}}Prec-Cart/Purchase\end{tabular}} & \textbf{\begin{tabular}[c]{@{}c@{}}F1-Cart/Purchase\end{tabular}} \\ \midrule
\multicolumn{1}{l|}{\textbf{OPT-2.7B}} & \multicolumn{1}{c|}{5253} & 0.74 & 0.869 & 0.917 & 0.55 & 0.402 & 0.551 & 0.532 & 0.255 & 0.533 & 0.542 \\ \midrule
\multicolumn{12}{l}{\textbf{OPT-350M Kmeans, K=6}} \\ \midrule
\multicolumn{1}{l|}{Cluster 1} & \multicolumn{1}{c|}{438} & 0.589 & 0.653 & 0.797 & 0.680 & 0.52 & 0.676 & 0.483 & 0.169 & 0.483 & 0.564 \\
\multicolumn{1}{l|}{Cluster 2} & \multicolumn{1}{c|}{1359} & 0.617 & 0.76 & 0.816 & 0.381 & 0.339 & 0.381 & 0.354 & 0.131 & 0.353 & 0.366 \\
\multicolumn{1}{l|}{Cluster 3} & \multicolumn{1}{c|}{638} & 0.633 & 0.82 & 0.884 & 0.545 & 0.2 & 0.545 & 0.339 & 0.05 & 0.339 & 0.418 \\
\multicolumn{1}{l|}{Cluster 4} & \multicolumn{1}{c|}{330} & 0.746 & 0.939 & 0.949 & 0.156 & 0.0 & 0.156 & 0.132 & 0.0 & 0.13 & 0.141 \\
\multicolumn{1}{l|}{Cluster 5} & \multicolumn{1}{c|}{573} & 0.745 & 0.944 & 0.951 & 0.248 & 0.238 & 0.248 & 0.333 & 0.238 & 0.333 & 0.284 \\
\multicolumn{1}{l|}{Cluster 6} & \multicolumn{1}{c|}{1915} & 0.688 & 0.786 & 0.858 & 0.509 & 0.479 & 0.509 & 0.486 & 0.201 & 0.488 & 0.498 \\
\multicolumn{1}{l|}{Combined} & \multicolumn{1}{c|}{5253} & 0.665 & 0.799 & 0.861 & 0.467 & 0.4 & 0.466 & 0.41 & 0.159 & 0.411 & 0.437 \\ \midrule
\multicolumn{12}{l}{\textbf{OPT-350M Kmeans, K=2}} \\ \midrule
\multicolumn{1}{l|}{Combined} & \multicolumn{1}{c|}{5253} & 0.698 & 0.831 & 0.884 & 0.467 & 0.376 & 0.466 & 0.458 & 0.184 & 0.458 & 0.462 \\ \midrule
\multicolumn{12}{l}{\textbf{OPT-350M \texttt{HeTLM} ($\alpha$=2, $\beta$=1)}} \\ \midrule
\multicolumn{1}{l|}{Cluster 1} & \multicolumn{1}{c|}{948} & 0.688 & 0.756 & 0.857 & 0.576 & 0.475 & 0.575 & 0.539 & 0.212 & 0.542 & 0.558 \\
\multicolumn{1}{l|}{Cluster 2} & \multicolumn{1}{c|}{319} & 0.865 & 0.966 & 0.969 & 0.136 & 0.1 & 0.136 & 0.545 & 0.333 & 0.545 & 0.218 \\
\multicolumn{1}{l|}{Cluster 3} & \multicolumn{1}{c|}{2875} & 0.745 & 0.865 & 0.917 & 0.530 & 0.391 & 0.529 & 0.577 & 0.267 & 0.577 & 0.552 \\
\multicolumn{1}{l|}{Cluster 4} & \multicolumn{1}{c|}{277} & 0.830 & 0.957 & 0.964 & 0.204 & 0.0 & 0.204 & 0.733 & 0.0 & 0.733 & 0.319 \\
\multicolumn{1}{l|}{Cluster 5} & \multicolumn{1}{c|}{759} & 0.696 & 0.80 & 0.862 & 0.459 & 0.417 & 0.459 & 0.364 & 0.103 & 0.364 & 0.407 \\
\multicolumn{1}{l|}{Combined} & \multicolumn{1}{c|}{5253} & 0.74 & 0.849 & 0.906 & 0.501 & 0.397 & 0.50 & 0.535 & 0.216 & 0.536 & 0.518 \\ \midrule
\multicolumn{12}{l}{\textbf{OPT-350M \texttt{HeTLM} ($\alpha$=5, $\beta$=9)}} \\ \midrule
\multicolumn{1}{l|}{Cluster 1} & \multicolumn{1}{c|}{305} & 0.872 & 0.974 & 0.974 & 0.073 & 0.0 & 0.073 & 0.75 & 0.0 & 0.75 & 0.133 \\
\multicolumn{1}{l|}{Cluster 2} & \multicolumn{1}{c|}{4797} & 0.728 & 0.859 & 0.912 & 0.529 & 0.383 & 0.528 & 0.536 & 0.245 & 0.536 & 0.532 \\
\multicolumn{1}{l|}{Cluster 3} & \multicolumn{1}{c|}{151} & 0.874 & 0.974 & 0.98 & 0.05 & 0.0 & 0.05 & 1.0 & 0.0 & 1.0 & 0.095 \\
\multicolumn{1}{l|}{Combined} & \multicolumn{1}{c|}{5253} & \textbf{0.741} & 0.869 & 0.918 & 0.51 & 0.371 & 0.508 & \textbf{0.537} & 0.245 & \textbf{0.537} & 0.522 \\ 
\midrule
\midrule
\midrule
\multicolumn{1}{l|}{\textbf{QWEN-2.5-7B}} & \multicolumn{1}{c|}{5253} & 0.674 & 0.877 & 0.916 & 0.394 & 0.147 & 0.394 & 0.411 & 0.153 & 0.411 & 0.403 \\ \midrule
\multicolumn{12}{l}{\textbf{QWEN-2.5-500M \texttt{HeTLM} ($\alpha$=5, $\beta$=9)}} \\ \midrule
\multicolumn{1}{l|}{Cluster 1} & \multicolumn{1}{c|}{3101} & 0.728 & 0.864 & 0.918 & 0.626 & 0.453 & 0.625 & 0.529 & 0.292 & 0.529 & 0.573 \\
\multicolumn{1}{l|}{Cluster 2} & \multicolumn{1}{c|}{316} & 0.88 & 0.975 & 0.978 & 0.273 & 0.111 & 0.273 & 0.667 & 1.0 & 0.667 & 0.387 \\
\multicolumn{1}{l|}{Cluster 3} & \multicolumn{1}{c|}{1673} & 0.725 & 0.885 & 0.917 & 0.549 & 0.475 & 0.548 & 0.530 & 0.301 & 0.530 & 0.539 \\
\multicolumn{1}{l|}{Cluster 4} & \multicolumn{1}{c|}{163} & 0.859 & 0.969 & 0.969 & 0.0 & 0.0 & 0.0 & 0.0 & 0.0 & 0.0 & 0.0 \\
\multicolumn{1}{l|}{Combined} & \multicolumn{1}{c|}{5253} & \textbf{0.74} & \textbf{0.881} & \textbf{0.923} & \textbf{0.58} & \textbf{0.446} & \textbf{0.579} & 0.531 & \textbf{0.296} & 0.531 & \textbf{0.554} \\ 
\bottomrule
\end{tabular}
}
\caption{Dataset I. Outcome Prediction Results. \textit{Higher is Better}. Bold indicates highest value in column. For \texttt{OPT}, \texttt{HeTLM} Combined results for $\alpha$=5, $\beta$=9, outperform OPT-350M-2.7B on 6 metrics, equals on 1, and is outperformed by 3. We find even stronger results favoring \texttt{HeTLM} for \texttt{QWEN2.5}.} 
\label{goog-data-step2-outcome-pred-compare}
\end{table*}

\subsection{Results: Step 1}
Small \texttt{LMs}, \texttt{OPT-350M}, \texttt{QWEN-2.5-500M} and \texttt{SmolLM2-360M}, are pretrained on Dataset I, while large \texttt{LMs}, \texttt{Llama3-8B}, \texttt{Mistral-7B} and \texttt{Gemma-7B}, are finetuned on the same data. Two variations of \texttt{Llama3-8B} are used - LORA finetuned and full finetuned. Per Table~\ref{tab:results-baseline-chatgpt}, small \texttt{LMs} outperform large \texttt{LMs}, namely, \texttt{GPT-4o-200B+}, \texttt{LLaMa2-7B}, \texttt{Llama3-8B}, \texttt{Mistral-7B}, \texttt{Gemma-7B}. Step 2 \texttt{HeTLM} clusterwise experiments are done with small \texttt{LMs}. 

\subsection{Results: Step 2}\label{sec:results-goog-step2}
Here we present results for \texttt{OPT} and \texttt{QWEN2.5}; \texttt{SmolLM2} results are in Appendix~\ref{subsec:exp_data1_smollm2}. In each table, the row \textbf{Combined} shows an average of all 5,253 users across clusters. First, see Table~\ref{goog-data-step2-page-gen-compare}. Comparative results are shown for \texttt{OPT-350M} and \texttt{QWEN-2.5-500M} because each has a large-sized \texttt{LM}, which is necessary for a just comparison, per Sec.~\ref{sec:choice-of-LM}. Since the clusterwise \texttt{HeTLM} for \texttt{OPT-350M} has $K=6$ times as many parameters (6*350M=2.1B), we compare against a larger, single \texttt{LM}, namely, \texttt{OPT-2.7B}. Similarly, we compare \texttt{HeTLM} for \texttt{QWEN-2.5-500M} with (6*500M=3.0B) parameters against larger, single \texttt{LM}, \texttt{QWEN-2.5-7B}. Results for \texttt{Kmeans} and \texttt{HeTLM} are shown. Varying $\alpha$, $\beta$ for \texttt{HeTLM} adjusts weights on losses in $\mathcal{L_O}$, leading to major changes in number of clusters and size. \texttt{OPT-350M-HeTLM} Combined ($\alpha$=5, $\beta$=9) outperforms all others and the baseline of single \texttt{LM OPT-2.7B} on all 6 metrics. For each combination of $\alpha$, $\beta$ (i) there is a large variation in evaluation metrics across clusters, implying that alignment with users varies across clusters, and (ii) cluster-specific metrics vary from the single \texttt{LM}, OPT-2.7B. 

Echoing results from Page Generation, in Outcome Prediction (Table~\ref{goog-data-step2-outcome-pred-compare}), \texttt{HeTLM} outperforms single \texttt{LM}, within each family \texttt{OPT} and \texttt{QWEN-2.5}. For Variance in Page Generation, Table~\ref{goog-data-step2-page-gen-variance} comparison of \texttt{HeTLM} $\alpha$=5, $\beta$=9 with respective \texttt{LM} shows for each combination of $\alpha$, $\beta$ (i) there is difference in variances across clusters for each metric, and (ii) cluster specific variances vary from the single \texttt{LLM OPT-2.7B}. The reduction in variance seen in Table~\ref{goog-data-step2-page-gen-variance} from \texttt{HeTLM}'s Combined evaluation supports the clustering approach. 

Table~\ref{goog-data-step2-comprehensive-eval} has a composite evaluation across all 20 metrics (see Sec.~\ref{sec:composie-metrics}). The overall Composite shows that respectively, 60\% and 80\% of times \texttt{HeTLM} for \texttt{OPT-350M} and \texttt{QWEN-2.5-500M} with $\alpha$=5, $\beta$=9 outperform the single, \texttt{LM} baselines, \texttt{OPT-2.7B} and \texttt{QWEN-2.5-7B}. For \texttt{OPT}, this is much higher than what exogenous clustering \texttt{Kmeans} achieves. In sum, page generation alignment vary across clusters and shows the downside of using a single \texttt{LM} to meet users' heterogeneous and subjective behaviors and preferences. \texttt{HeTLM} offers better alignment with users. 

In addition to results shown for Kmeans, K=6, we also show results for K=2 (decided based on Silhouette Coefficient). See Tables~\ref{goog-data-step2-page-gen-compare} -~\ref{goog-data-step2-comprehensive-eval}. We find that K=2 produces worse performance than K=6 for \texttt{Kmeans}, where choice of number of clusters is fixed exogenously. However, notably for \texttt{HeTLM}, choice of number of clusters is endogenous and \texttt{HeTLM} outperforms \texttt{Kmeans}.

\begin{table}[hbt]
\resizebox{\columnwidth}{!}{ 
\begin{tabular}{l|cccc}
\toprule
\textbf{Model} & \textbf{\begin{tabular}[c]{@{}c@{}}Outcome \\ Prediction\end{tabular}} & \textbf{\begin{tabular}[c]{@{}c@{}}Page Gen\\ Mean\end{tabular}} & \textbf{\begin{tabular}[c]{@{}c@{}}Page Gen\\ Var\end{tabular}} & \textbf{\begin{tabular}[c]{@{}c@{}}Overall \\ Composite\end{tabular}} \\
\midrule
\multicolumn{5}{l}{\textbf{Compared with OPT-2.7B}} \\ \midrule
\multicolumn{5}{l}{\textbf{OPT-350M - Combined}} \\ \midrule 
\textbf{Kmeans, K=6} & 0.0   & 0.333 & 0.75  & 0.25 \\
\textbf{Kmeans, K=2} & 0.1 & 0.0 & 0.5 & 0.15 \\
\textbf{\texttt{HeTLM} ($\alpha$=2, $\beta$=1)} & 0.3   & 0.167 & 0.75   & 0.35 \\
\textbf{\texttt{HeTLM} ($\alpha$=5, $\beta$=9)} & 0.5   & 0.5   & 1.0  & 0.6 \\
\midrule
\midrule
\midrule
\multicolumn{5}{l}{\textbf{Compared with QWEN-2.5-7B}} \\ \midrule
\multicolumn{5}{l}{\textbf{QWEN-2.5-500M - Combined}} \\ \midrule 
\textbf{\texttt{HeTLM} ($\alpha$=5, $\beta$=9)} & 1.0   & 0.833   & 0.25  & 0.8 \\
\bottomrule
\end{tabular}
}
\caption{Dataset I. Composite metrics of Sec.~\ref{sec:composie-metrics}. 
  Overall Composite clearly favors \texttt{HeTLM} $\alpha$=5, $\beta$=9 over other variation and  exogenous clusterwise \texttt{Kmeans}.}
\label{goog-data-step2-comprehensive-eval}
\end{table}

\begin{table}[bth]
\resizebox{\columnwidth}{!}{%
    \begin{tabular}{@{}l|cccc@{}} 
    \toprule
    \multicolumn{1}{c|}{\textbf{Model}} & \textbf{BLEU} & \textbf{BERT-P} & \textbf{BERT-R} & \textbf{BERT-F1} \\ \midrule
    \textbf{LLaMa-3-8B} & 0.253 & 0.891 & 0.889 & 0.889 \\ \midrule
    \textbf{Mistral-7B} & 0.196 & 0.888 & 0.88 & 0.883 \\ \midrule
    \textbf{Gemma-7B} & 0.255 & 0.889 & 0.88 & 0.889 \\ \midrule
    \midrule
    \midrule
    \multicolumn{5}{@{}l}{\textbf{Small-LMs}} \\ \midrule
    \textbf{OPT-350M} & 0.320 & 0.903 & 0.896 & 0.899 \\ \midrule
    \textbf{SmolLM2-360M} & 0.360 & 0.906 & 0.903 & 0.904 \\ \midrule
    \textbf{QWEN-2.5-500M} & 0.355 & 0.903 & 0.902 & 0.902 \\
    \midrule
    \midrule
    \midrule
    \textbf{OPT 2.7B} & 0.376 & 0.906 & 0.908 & 0.907 \\ \midrule
    \multicolumn{5}{@{}l}{\textbf{\texttt{OPT-350M Kmeans}, K=6}} \\ \midrule
    Cluster 1 & 0.241 & 0.861 & 0.889 & 0.874 \\
    Cluster 2 & 0.137 & 0.814 & 0.844 & 0.828 \\
    Cluster 3 & 0.20 & 0.855 & 0.878 & 0.866 \\
    Cluster 4 & 0.39 & 0.903 & 0.906 & 0.904 \\
    Cluster 5 & 0.305 & 0.917 & 0.890 & 0.903 \\
    Cluster 6 & 0.268 & 0.891 & 0.888 & 0.888 \\
    Combined & 0.235 & 0.868 & 0.878 & 0.871 \\ \midrule
    \multicolumn{5}{@{}l}{\textbf{\texttt{OPT-350M HeTLM} ($\alpha$=5, $\beta$=9)}} \\ \midrule
    Cluster 1 & \underline{0.637} & \underline{0.954} & \underline{0.944} & \underline{0.949} \\
    Cluster 2 & 0.339 & 0.902 & 0.901 & 0.901 \\ 
    Cluster 3 & 0.403 & 0.951 & 0.895 & 0.922 \\
    Combined & 0.358 & 0.907 & 0.904 & 0.905 \\ \midrule
    \midrule
    \midrule
    \textbf{QWEN-2.5-7B} & 0.254 & 0.892 & 0.888 & 0.889 \\ \midrule
    \multicolumn{5}{@{}l}{\textbf{\texttt{QWEN-2.5-500M HeTLM} ($\alpha$=5, $\beta$=9)}} \\ \midrule
    Cluster 1 & 0.359 & 0.907 & 0.907 & 0.907 \\
    Cluster 2 & \underline{0.63} & \underline{0.954} & \underline{0.945} & \underline{0.948} \\
    Cluster 3 & 0.365 & 0.911 & 0.903 & 0.906 \\
    Cluster 4 & 0.396 & 0.940 & 0.898 & 0.918 \\
    Combined & \textbf{0.378} & \textbf{0.912} & \textbf{0.908} & \textbf{0.909} \\ \bottomrule
    \end{tabular}%
}
\caption{Dataset I. BLEU and BERT scores. Bold indicates the highest value in the column. Underline shows comparison of the larger \texttt{LM} with \texttt{HeTLM} for small \texttt{LM} of the same family. 
}
\label{goog-data-bleu-bert-compare}
\end{table}

\subsection{BLEU and BERT metrics} 
See Table~\ref{goog-data-bleu-bert-compare}. Remarkably, each small, single \texttt{LM} outperforms each large, single \texttt{LM} in BLEU and in all 3 BERT metrics. Within \texttt{QWEN-2.5}, \texttt{QWEN2.5-HeTLM} Combined performs better than the single \texttt{LM}, \texttt{QWEN-2.5 7B}, in all metrics, and also performs better than all \texttt{LMs} shown. The improvements at cluster level (Cluster 2 for \texttt{QWEN-2.5-HeTLM}, Cluster 1 for \texttt{OPT-350M-HeTLM}) and differences across clusters in the metrics are notable. The results emphasize the importance of recognizing heterogeneity in users' subjective behaviors when training \texttt{LMs} so that better alignment ensues across users.

\begin{table}[t]
\resizebox{\columnwidth}{!}{ 
\begin{tabular}{l|cccc}
\toprule
\textbf{Model} & \textbf{\begin{tabular}[c]{@{}c@{}}Outcome \\ Prediction\end{tabular}} & \textbf{\begin{tabular}[c]{@{}c@{}}Page Gen\\ Mean\end{tabular}} & \textbf{\begin{tabular}[c]{@{}c@{}}Page Gen\\ Var\end{tabular}} & \textbf{\begin{tabular}[c]{@{}c@{}}Overall \\ Composite\end{tabular}} \\
\midrule
\multicolumn{5}{l}{\textbf{Compared with OPT-2.7B}} \\ \midrule
\multicolumn{5}{l}{\textbf{OPT-350M - Combined}} \\ \midrule 
\textbf{\texttt{HeTLM} ($\alpha$=1, $\beta$=0)} & 0.1   & 0.0 & 0.75   & 0.2 \\
\textbf{\texttt{HeTLM} ($\alpha$=1, $\beta$=3)} & 0.4   & 0.167 & 0.75  & 0.4 \\  
\textbf{\texttt{HeTLM} ($\alpha$=9, $\beta$=0)} & 0.1   & 0.167 & 0.75   & 0.25 \\
\textbf{\texttt{HeTLM} ($\alpha$=9, $\beta$=3)} & 0.1   & 0.0   & 0.75  & 0.2 \\
\bottomrule
\end{tabular}
}
\caption{Dataset I. Ablation. Composite metrics of Sec.~\ref{sec:composie-metrics}. Ablation with \texttt{OPT-350M, Pre-training}, by varying $\alpha$, $\beta$.}
\label{goog-data-opt-350m-ablation-step2-comprehensive-eval}
\end{table}

\subsection{Ablation: \texttt{OPT-350M}}
We ran two ablations to examine the effect of the sharpness of the selector probability distribution ($\alpha$) and the presence of cluster separation ($\beta$). The hyperparameters $\alpha$, $\beta$ are selected empirically and these are dataset and \texttt{LM} specific. We provide a snapshot of composite metrics in Table~\ref{goog-data-opt-350m-ablation-step2-comprehensive-eval}. By comparing with results of \texttt{HeTLM} for \texttt{OPT-350M} in Table~\ref{goog-data-step2-comprehensive-eval}, we find that the proposed architecture with non-zero $\alpha$=5, $\beta$=9 performs appreciably better than suppressing either $\alpha$, or $\beta$. General trends may be identified with a more indepth analysis which is beyond the scope of this paper.

\section{Conclusion and Discussion}
We address a specific research gap in the otherwise vast and growing literature in \texttt{LLMs}. This gap emanates from lack of attention to the language of browsing, which is idiosyncratically generated as sequences of pages, by each user as s/he subjectively browses websites or apps. This language does not have the grammar and structure of natural language. Working with the language of browsing, users' heterogeneity of behaviors and subjective preferences call for a model with endogenous clusterwise training to balance between performances on page generation, outcome prediction and reducing variance in alignment. While training an \texttt{LM} satisfying these objectives is the primary goal, as way of applications, we propose using the \texttt{LM} thus trained to derive solutions to make a variety of everyday business tasks \textit{predictive}. These tasks range from predictive targeting (based on product page in next session), predictive journey (based on sequence of pages in next session), segmentation (based on predictive journey) and recommendation. Future work can overcome some limitations of this paper by delving into larger browsing datasets and these predictive tasks, and compare with conventional models. Also, \texttt{HeTLM} shows lower inference time than a relevant single \texttt{LM} (Appendix Table~\ref{tab:inference-times}). 

\section*{Limitations}

There are a few limitations to which we draw attention. For context, the language of browsing is commonplace since every online firm collects behavior logs of sequences of pages every user clicks on its website or app. Yet, unlike natural language data that are readily available on the internet, this type of data is private to the firm and reside in a protected data lake. We obtained two datasets put out in the public domain and show experiments with those. We tried to make the most of these datasets by performing experiments on 3 small \texttt{LMs} and 6 large \texttt{LMs} to provide a fair comparison. 

As limitations: One, to generalize our findings, it will be valuable to run our proposed \texttt{HeTLM} on other language of browsing datasets. Two, due to the fundamental differences between browsing sequences and natural language, standard NLP datasets could not be used for evaluation. Three, a scalability analysis, out of scope for the paper and given our access to limited compute, will be useful going forward. Four, staying with single \texttt{LMs}, we compare small versus large \texttt{LMs} to show better performance of small \texttt{LMs}, where the large \texttt{LMs} up to 8B in size are finetuned on our data. Extending this to much larger sized \texttt{LMs}, such as Llama-70B or others which are finetunable, will provide additional test of our proposition of using small \texttt{LMs} for language of browsing. 

\bibliography{custom}

\onecolumn
\newpage
\twocolumn
\appendix

\section{Appendix}
\label{sec:appendix}

\subsection{Theoretical Basis of \texttt{HeTLM}} \label{sec:theoretical-basis}

Following \cite{lee2020temporalphenotypingusingdeep}, we formulate the problem of identifying user clusters with similar browsing behaviors and preferences as a predictive clustering problem. Let $X \in \mathcal{X}$ and $Y \in \mathcal{Y}$ be random variables for input browsing sessions and output next session pages with a joint distribution $p_{XY}$, where $\mathcal{X}$ is the input session space and $\mathcal{Y}$ is the next session page space.

For each user $n$, we are given sequences of browsing sessions $x^{n}$. Our aim is to identify a set of $K$ predictive clusters, $\mathcal{C} = \{\mathcal{C}(1), ..., \mathcal{C}(K)\}$, where each cluster consists of users with similar browsing behaviors and preferences.

We define a cluster as $\mathcal{C}(k) = \{x^{n}|t,~s^{n}=k\}$ for $k\in \{1,...,K\}$ where $s^{n} \in \{1,...,K\}$ is the cluster assignment for user $n$. This allows us to flexibly update the cluster assignment to which a user belongs as the cluster representations are updated over time.

Let $S$ be a random variable for the cluster assignment. We want to find an optimal partitioning of users into $K$ clusters such that the difference in the next session distribution conditioned on the input sessions $X$ and the cluster assignment $S$ is minimized, while optimizing the number of clusters $K$. This can be achieved by minimizing the following Kullback-Leibler (KL) divergence:
\begin{equation}
\label{eq:theoretical_kl}
\underset{K}{\text{minimize}} \sum_{k=1}^{K} \sum_{x\in\mathcal{C}(k)} KL(Y|X=x \| Y|S=k)
\end{equation}

The optimization problem is non-trivial, and we estimate this objective function through the LLM loss ($\mathcal{L}_1$) described in Sec. \ref{sec:all-losses}. Minimizing $\mathcal{L}_1$ is equivalent to minimizing the KL divergence in Eq. \ref{eq:theoretical_kl} since the former term is independent of our optimization procedure. To avoid trivial solutions in this unsupervised setting, such as all embeddings collapsing to a single point or the selector assigning equal probability to all clusters regardless of the input sequence, we introduce two auxiliary loss functions, $\mathcal{L}_2$ and $\mathcal{L}_3$, as detailed in Sec. \ref{sec:all-losses}. 

\subsection{Experimental Results: \texttt{HeTLM} for \texttt{SmolLM2-360M}}
\label{subsec:exp_data1_smollm2}
For small \texttt{LM}, \texttt{SmolLM2-360M}, we present results for clusterwise training. The clusterwise \texttt{HeTLM} results here show considerable variations across clusters in all metrics of  Tables~\ref{google-data-smollm2-page-gen-mean}, \ref{google-data-smollm2-page-gen-variance}, and \ref{google-data-smollm2-outcome-prediction}. This calls attention to the importance of clusterwise training to achieve better alignment of users' behaviors.

\begin{table}[t]
  
\resizebox{\columnwidth}{!}{ 
\begin{tabular}{@{}lccccccc@{}}
\toprule
\multicolumn{1}{l|}{\textbf{Model}} & \multicolumn{1}{c|}{\textbf{$N$}} & \textbf{\begin{tabular}[c]{@{}c@{}}HR\end{tabular}} & \textbf{\begin{tabular}[c]{@{}c@{}}IoA\end{tabular}} & \textbf{\begin{tabular}[c]{@{}c@{}}IoP\end{tabular}} & \textbf{\begin{tabular}[c]{@{}c@{}}IoU\end{tabular}} & \textbf{\begin{tabular}[c]{@{}c@{}}New-P\end{tabular}} & \textbf{\begin{tabular}[c]{@{}c@{}}Val-P\end{tabular}} \\ \midrule
\multicolumn{8}{l}{\textbf{\texttt{SmolLM2-360M: HeTLM} ($\alpha$=5, $\beta$=9)}} \\ \midrule
\multicolumn{1}{l|}{Cluster 1} & \multicolumn{1}{c|}{296} & 0.889 & 0.661 & 0.585 & 0.493 & 0.014 & 0.002 \\
\multicolumn{1}{l|}{Cluster 2} & \multicolumn{1}{c|}{4806} & 0.809 & 0.419 & 0.348 & 0.253 & 0.262 & 0.05 \\
\multicolumn{1}{l|}{Cluster 3} & \multicolumn{1}{c|}{151} & 0.570 & 0.310 & 0.439 & 0.29 & 0.0 & 0.0 \\
\multicolumn{1}{l|}{Combined} & \multicolumn{1}{c|}{5253} & 0.806 & 0.429 & 0.364 & 0.268 & 0.24 & 0.046 \\ \bottomrule
\end{tabular}
}
\caption{Dataset I. Page Generation Results. \textit{Higher is better}. One \texttt{HeTLM} for \texttt{SmolLM2-360M} version is shown.}
\label{google-data-smollm2-page-gen-mean}
\end{table}

\begin{table}[h]
  
\resizebox{\columnwidth}{!}{%
\begin{tabular}{@{}lccccc@{}}
\toprule
\multicolumn{1}{l|}{\textbf{Model}} & \multicolumn{1}{c|}{\textbf{$N$}} & \textbf{\begin{tabular}[c]{@{}c@{}}HR-var\end{tabular}} & \textbf{\begin{tabular}[c]{@{}c@{}}IoA-var\end{tabular}} & \textbf{\begin{tabular}[c]{@{}c@{}}IoP-var\end{tabular}} & \textbf{\begin{tabular}[c]{@{}c@{}}IoU-var\end{tabular}} \\ \midrule
\multicolumn{6}{l}{\textbf{\texttt{SmolLM2-360M: HeTLM} ($\alpha$=5, $\beta$=9)}} \\ \midrule
\multicolumn{1}{l|}{Cluster 1} & \multicolumn{1}{c|}{296} & 0.099 & 0.125 & 0.091 & 0.09 \\
\multicolumn{1}{l|}{Cluster 2} & \multicolumn{1}{c|}{4806} & 0.155 & 0.113 & 0.082 & 0.068 \\
\multicolumn{1}{l|}{Cluster 3} & \multicolumn{1}{c|}{151} & 0.245 & 0.145 & 0.187 & 0.142 \\
\multicolumn{1}{l|}{Combined} & \multicolumn{1}{c|}{5253} & 0.156 & 0.118 & 0.088 & 0.075 \\ \bottomrule
\end{tabular}
}
\caption{Dataset I. Variance in Page Generation across users. \textit{Lower is better}. One \texttt{HeTLM} for \texttt{SmolLM2-360M} version is shown.}
\label{google-data-smollm2-page-gen-variance}
\end{table}

\begin{table*}[h]
  
\resizebox{\textwidth}{!}{  
\begin{tabular}{@{}lccccccccccc@{}}
\toprule
\multicolumn{1}{l|}{\textbf{Model}} & \multicolumn{1}{c|}{\textbf{$N$}} & \textbf{Cart Acc} & \textbf{\begin{tabular}[c]{@{}c@{}}Acc-Purchase\end{tabular}} & \textbf{\begin{tabular}[c]{@{}c@{}}Acc-Cart/Purchase\end{tabular}} & \textbf{\begin{tabular}[c]{@{}c@{}}Rec-Cart\end{tabular}} & \textbf{\begin{tabular}[c]{@{}c@{}}Rec-Purchase\end{tabular}} & \textbf{\begin{tabular}[c]{@{}c@{}}Rec-Cart/Purchase\end{tabular}} & \textbf{\begin{tabular}[c]{@{}c@{}}Prec-Cart\end{tabular}} & \textbf{\begin{tabular}[c]{@{}c@{}}Prec-Purchase\end{tabular}} & \textbf{\begin{tabular}[c]{@{}c@{}}Prec-Cart/Purchase\end{tabular}} & \textbf{\begin{tabular}[c]{@{}c@{}}F1-Cart/Purchase\end{tabular}} \\ \midrule
\multicolumn{12}{l}{\textbf{\texttt{SmolLM2-360M: HeTLM} ($\alpha$=5, $\beta$=9)}} \\ \midrule
\multicolumn{1}{l|}{Cluster 1} & \multicolumn{1}{c|}{296} & 0.868 & 0.97 & 0.976 & 0.225 & 0.0 & 0.225 & 0.529 & 0.0 & 0.529 & 0.316 \\
\multicolumn{1}{l|}{Cluster 2} & \multicolumn{1}{c|}{4806} & 0.710 & 0.845 & 0.902 & 0.609 & 0.489 & 0.608 & 0.503 & 0.249 & 0.504 & 0.511 \\
\multicolumn{1}{l|}{Cluster 3} & \multicolumn{1}{c|}{151} & 0.868 & 0.974 & 0.974 & 0.05 & 0.0 & 0.05 & 0.5 & 0.0 & 0.5 & 0.091 \\
\multicolumn{1}{l|}{Combined} & \multicolumn{1}{c|}{5253} & 0.724 & 0.856 & 0.908 & 0.591 & 0.474 & 0.590 & 0.504 & 0.249 & 0.504 & 0.544 \\ \bottomrule
\end{tabular}
}
\caption{Dataset I. Outcome Prediction Results \textit{Higher is better}. \texttt{HeTLM} for \texttt{SmolLM2-360M}}
\label{google-data-smollm2-outcome-prediction}
\end{table*}

\begin{table*}[h]
\resizebox{\textwidth}{!}{  
\begin{tabular}{@{}lccccccccccc@{}}
\toprule
\multicolumn{1}{l|}{\textbf{Model}} & \multicolumn{1}{c|}{\textbf{$N$}} & \textbf{Cart Acc} & \textbf{\begin{tabular}[c]{@{}c@{}}Acc-Purchase\end{tabular}} & \textbf{\begin{tabular}[c]{@{}c@{}}Acc-Cart/Purchase\end{tabular}} & \textbf{\begin{tabular}[c]{@{}c@{}}Rec-Cart\end{tabular}} & \textbf{\begin{tabular}[c]{@{}c@{}}Rec-Purchase\end{tabular}} & \textbf{\begin{tabular}[c]{@{}c@{}}Rec-Cart/Purchase\end{tabular}} & \textbf{\begin{tabular}[c]{@{}c@{}}Prec-Cart\end{tabular}} & \textbf{\begin{tabular}[c]{@{}c@{}}Prec-Purchase\end{tabular}} & \textbf{\begin{tabular}[c]{@{}c@{}}Prec-Cart/Purchase\end{tabular}} & \textbf{\begin{tabular}[c]{@{}c@{}}F1-Cart/Purchase\end{tabular}} \\ \midrule
\multicolumn{12}{l}{\textbf{Single \texttt{LLM}}} \\ \midrule
\multicolumn{1}{l|}{OPT 2.7B} & \multicolumn{1}{c|}{10000} & 0.689 & 0.874 & 0.935 & 0.629 & 0.504 & 0.627 & 0.475 & 0.341 & 0.482 & 0.545 \\ \midrule
\multicolumn{12}{l}{\textbf{Kmeans, K=2}} \\ \midrule
\multicolumn{1}{l|}{Cluster 1} & \multicolumn{1}{c|}{5421} & 0.716 & 0.904 & 0.94 & 0.452 & 0.401 & 0.451 & 0.448 & 0.313 & 0.454 & 0.453 \\
\multicolumn{1}{l|}{Cluster 2} & \multicolumn{1}{c|}{4579} & 0.666 & 0.857 & 0.942 & 0.76 & 0.49 & 0.757 & 0.499 & 0.384 & 0.508 & 0.608 \\
\multicolumn{1}{l|}{Combined} & \multicolumn{1}{c|}{10000} & 0.693 & 0.883 & 0.941 & 0.612 & 0.453 & 0.612 & 0.48 & 0.355 & 0.488 & 0.543 \\ \midrule
\multicolumn{12}{l}{\textbf{\texttt{HeTLM} ($\alpha$=5, $\beta$=9)}} \\ \midrule
\multicolumn{1}{l|}{Cluster 1} & \multicolumn{1}{c|}{121} & 0.587 & 0.835 & 0.917 & 0.617 & 0.143 & 0.604 & 0.475 & 0.2 & 0.475 & 0.532 \\
\multicolumn{1}{l|}{Cluster 2} & \multicolumn{1}{c|}{9879} & 0.706 & 0.875 & 0.933 & 0.578 & 0.514 & 0.578 & 0.496 & 0.344 & 0.504 & 0.538 \\
\multicolumn{1}{l|}{Combined} & \multicolumn{1}{c|}{10000} & 0.705 & 0.875 & 0.933 & 0.578 & 0.508 & 0.578 & 0.496 & 0.343 & 0.504 & 0.538 \\ \bottomrule
\end{tabular}
}
\caption{Dataset II. Outcome Prediction Results \textit{Higher is better}. Top performing versions of \texttt{Kmeans} and \texttt{HeTLM} are shown. Results of other versions for Dataset II are in Appendix.}
\label{ecommerce-data-outcome-prediction}
\end{table*}

\subsection{Dataset II: Experiments}
\label{subsec:exp_data2}
This dataset is public as well~\cite{mkechinov_ecommerce_behavior_2019}. Train : Test split = 1M : 10,000 samples. Number of unique pages = 17,310. The dataset has millions of consumers; after applying filters for excluding very short sessions and incomplete data, we randomly select 1.01M samples. \textbf{Outcomes} available in the pageurls are: \textit{Cart} (Add to Cart, in Dataset II) and \textit{Purchase}, these Outcome pages are generated by the \texttt{LM} and also used as \textit{target labels} to evaluate Outcome prediction. 

\subsubsection{Experimental Results: Dataset II}

We present comparison of \texttt{HeTLM} with \texttt{OPT-2.7B}.
In Page Generation, comparing the respective Combined row in Table~\ref{ecommerce-data-page-gen-mean} with \texttt{OPT-2.7B} shows that except Val-P, in all other metrics \texttt{Kmeans} and \texttt{HeTLM} perform slightly better. For \texttt{HeTLM}, the variations in number of clusters and size of clusters and their differences in metrics justify endogenous clustering based training. In Outcome Prediction (Table~\ref{ecommerce-data-outcome-prediction}) across its 10 metrics \texttt{HeTLM} has an edge over \texttt{OPT-2.7B}. For Variance in Page Generation (Table~\ref{ecommerce-data-page-gen-variance}) both models perform similarly. Reviewing Overall Composite metric in Table~\ref{ecommerce-data-composite-eval} finds \texttt{HeTLM} outperforming \texttt{OPT-2.7B}, as the values [0.7, 0.6] are $>$ 0.5. APPENDIX contains additional results from Dataset II for more variations of \texttt{HeTLM} and \texttt{Kmeans}.

\begin{table}[h]
  
\resizebox{\columnwidth}{!}{ 
\begin{tabular}{@{}lccccccc@{}}
\toprule
\multicolumn{1}{l|}{\textbf{Model}} & \multicolumn{1}{c|}{\textbf{$N$}} & \textbf{\begin{tabular}[c]{@{}c@{}}HR\end{tabular}} & \textbf{\begin{tabular}[c]{@{}c@{}}IoA\end{tabular}} & \textbf{\begin{tabular}[c]{@{}c@{}}IoP\end{tabular}} & \textbf{\begin{tabular}[c]{@{}c@{}}IoU\end{tabular}} & \textbf{\begin{tabular}[c]{@{}c@{}}New-P\end{tabular}} & \textbf{\begin{tabular}[c]{@{}c@{}}Val-P\end{tabular}} \\ \midrule
\multicolumn{1}{l|}{OPT 2.7B} & \multicolumn{1}{c|}{10000} & 0.542 & 0.232 & 0.311 & 0.186 & 1.0 & 0.917 \\ \midrule
\multicolumn{8}{l}{\textbf{Kmeans, K=2}} \\ \midrule
\multicolumn{1}{l|}{Cluster 1} & \multicolumn{1}{c|}{5421} & 0.462 & 0.171 & 0.25 & 0.136 & 1.0 & 0.932 \\
\multicolumn{1}{l|}{Cluster 2} & \multicolumn{1}{c|}{4579} & 0.642 & 0.304 & 0.392 & 0.248 & 1.0 & 0.897 \\
\multicolumn{1}{l|}{Combined} & \multicolumn{1}{c|}{10000} & 0.544 & 0.232 & 0.315 & 0.187 & 1.0 & 0.916 \\ \midrule
\multicolumn{8}{l}{\textbf{\texttt{HeTLM} ($\alpha$=5, $\beta$=9)}} \\ \midrule
\multicolumn{1}{l|}{Cluster 1} & \multicolumn{1}{c|}{121} & 0.512 & 0.215 & 0.2 & 0.136 & 0.983 & 0.672 \\
\multicolumn{1}{l|}{Cluster 2} & \multicolumn{1}{c|}{9879} & 0.545 & 0.236 & 0.312 & 0.188 & 1.0 & 0.904 \\
\multicolumn{1}{l|}{Combined} & \multicolumn{1}{c|}{10000} & 0.544 & 0.236 & 0.311 & 0.188 & 1.0 & 0.901 \\ \bottomrule
\end{tabular}
}
\caption{Dataset II. Page Generation Results. \textit{Higher is better}. Top version of \texttt{Kmeans} model, top  \texttt{HeTLM} versions are shown.}
\label{ecommerce-data-page-gen-mean}
\end{table}

\begin{table}[h]
\resizebox{\columnwidth}{!}{%
\begin{tabular}{@{}lccccc@{}}
\toprule
\multicolumn{1}{l|}{\textbf{Model}} & \multicolumn{1}{c|}{\textbf{$N$}} & \textbf{\begin{tabular}[c]{@{}c@{}}HR-var\end{tabular}} & \textbf{\begin{tabular}[c]{@{}c@{}}IoA-var\end{tabular}} & \textbf{\begin{tabular}[c]{@{}c@{}}IoP-var\end{tabular}} & \textbf{\begin{tabular}[c]{@{}c@{}}IoU-var\end{tabular}} \\ \midrule
\multicolumn{1}{l|}{OPT 2.7B} & \multicolumn{1}{c|}{10000} & 0.248 & 0.077 & 0.118 & 0.058 \\ \midrule
\multicolumn{6}{l}{\textbf{Kmeans, K=2}} \\ \midrule
\multicolumn{1}{l|}{Cluster 1} & \multicolumn{1}{c|}{5421} & 0.249 & 0.057 & 0.103 & 0.042 \\
\multicolumn{1}{l|}{Cluster 2} & \multicolumn{1}{c|}{4579} & 0.23 & 0.089 & 0.129 & 0.071 \\
\multicolumn{1}{l|}{Combined} & \multicolumn{1}{c|}{10000} & 0.248 & 0.076 & 0.12 & 0.059 \\ \midrule
\multicolumn{6}{l}{\textbf{\texttt{HeTLM} ($\alpha$=5, $\beta$=9)}} \\ \midrule
\multicolumn{1}{l|}{Cluster 1} & \multicolumn{1}{c|}{121} & 0.25 & 0.072 & 0.057 & 0.031 \\
\multicolumn{1}{l|}{Cluster 2} & \multicolumn{1}{c|}{9879} & 0.248 & 0.079 & 0.117 & 0.059 \\
\multicolumn{1}{l|}{Combined} & \multicolumn{1}{c|}{10000} & 0.248 & 0.079 & 0.116 & 0.058 \\ \bottomrule
\end{tabular}
}
\caption{Dataset II. Variance in Page Generation across users \textit{lower is better}.}
\label{ecommerce-data-page-gen-variance}
\end{table}

\begin{table}[h]
\resizebox{\columnwidth}{!}{ 
\begin{tabular}{l|cccc}
\toprule
\textbf{Model} & \textbf{\begin{tabular}[c]{@{}c@{}}Outcome \\ Pred\end{tabular}} & \textbf{\begin{tabular}[c]{@{}c@{}}Page Gen\\ Mean\end{tabular}} & \textbf{\begin{tabular}[c]{@{}c@{}}Page Gen\\ Var\end{tabular}} & \textbf{\begin{tabular}[c]{@{}c@{}}Overall \\ Composite\end{tabular}} \\
\midrule
\textbf{\texttt{Kmeans}, K=2} & 0.6 & 0.5 & 0.5 & 0.55 \\
\textbf{\texttt{HeTLM} ($\alpha$=5, $\beta$=9)} & 0.6 & 0.667 & 0.5 & 0.6 \\
\bottomrule
\end{tabular}
}
\caption{Dataset II. Composite metrics.  
  \texttt{HeTLM} outperforms \texttt{OPT-2.7B} and \texttt{Kmeans }in 3 metrics, and equals in Page Generation - Variance.
  } 
\label{ecommerce-data-composite-eval}
\end{table}

\subsection{Experimental Setup Details}
\label{subsec:exp_setup}

All our experiments are performed on EC2 p4de.24xlarge~\cite{AmazonEC67:online} instances with 8 A100 GPUs~\cite{9361255} with each having 80 GB GPU (HBM) memory. For fine-tuning we used the open-source \texttt{Axolotl} library~\cite{axolotl2023}. We provide the hyperparameters used in our experiments for small model pretraining, heterogeneity-aware model training, and large model fine-tuning below.

\begin{table}[t!]
\centering

\resizebox{\columnwidth}{!}{%
\begin{tabular}{lc}
\toprule
\textbf{Hyperparameter} & \textbf{Value} \\
\midrule
\multicolumn{2}{c}{\textit{General Configuration}} \\
\midrule
Base Model & \begin{tabular}[c]{@{}c@{}}OPT-350M (Pretrained)\\QWEN-2.5-500M (Pretrained)\\SmolLM2-360M (Pretrained)\end{tabular} \\
Tokenizer Type & Custom (page level) \\
Batch Size & 16 \\
Maximum Sequence Length & 512 \\
Validation Set Size & 0.05 \\
Number of Epochs & 10 \\
Number of Clusters (Initial) & 6 \\
\midrule
\multicolumn{2}{c}{\textit{Optimizer Settings}} \\
\midrule
Selector Learning Rate & 5e-5 \\
Predictor Learning Rate & 1e-5 \\
Selector Weight Decay & 1e-4 \\
Predictor Weight Decay & 0.01 \\
\midrule
\multicolumn{2}{c}{\textit{Loss Weights}} \\
\midrule
$\alpha$ (Loss $\mathcal{L}_2$ weight) & 0.5--5.0 \\
$\beta$ (Loss $\mathcal{L}_3$ weight) & 1.0--9.0 \\
\midrule
\multicolumn{2}{c}{\textit{Encoder Configuration}} \\
\midrule
Encoder Type & Sentence Transformers \\
Encoder Path & all-MiniLM-L6-v2 \\
\bottomrule
\end{tabular}
}
\caption{HeTLM Training Hyperparameters}
\label{tab:hetlm-training}
\end{table}

\subsubsection{Prompt Templates}

The zero-shot setup used a basic instruction prompt: "A user's website browsing sequence of pages over multiple sessions is given as Input. {INPUT SESSIONS} [BOS] denotes beginning of session, [EOS] denotes end of session. Based on activity in the given input sessions, predict the next session as a sequence of pages highly likely to be visited by the user". For the few-shot setup we add 3 examples of input-output pairs based on cosine similarity of SBERT embeddings with the user's input sessions: "Learn from similar users browsing sequences given below: {INPUT-OUTPUT PAIRS}".

\subsubsection{Small Model Pretraining}
\label{subsec:small-model-hp}

Table~\ref{tab:small-model-pretraining} presents the hyperparameters for pretraining the small \texttt{LM}. This model was pretrained on browsing session data before being used in other experiments.

\begin{table}[t!]
\centering
\resizebox{\columnwidth}{!}{%
\begin{tabular}{lc}
\toprule
\textbf{Hyperparameter} & \textbf{Value} \\
\midrule
Base Model & \begin{tabular}[c]{@{}c@{}}OPT-350M\\QWEN-2.5-500M\\SmolLM2-360M\end{tabular} \\
Tokenizer Type & Custom (page level) \\
Learning Rate & 1e-5 \\
Weight Decay & 0.01 \\
Batch Size & 32 \\
Maximum Sequence Length & 512 \\
Validation Set Size & 0.05 \\
Number of Epochs & 100 \\
Gradient Accumulation Steps & 4 \\
Warmup Ratio & 0.05 \\
\bottomrule
\end{tabular}
}
\caption{Small \texttt{LM} Pretraining Hyperparameters}
\label{tab:small-model-pretraining}
\end{table}

\subsubsection{HeTLM Training}
\label{subsec:hetlm-hp}

Table~\ref{tab:hetlm-training} details the hyperparameters used for training our proposed Heterogeneity-aware Training of Language Models (HeTLM). These parameters control the Actor-Critic architecture and the weight of various loss components.

\subsubsection{Large Model LoRA Finetuning}
\label{subsec:lora-hp}

Table~\ref{lora-finetuning} lists the hyperparameters used for finetuning large \texttt{LMs} using Low-Rank Adaptation (LoRA). These settings were applied across different model architectures for comparison purposes.

\begin{table}[t!]
\centering

\label{tab:lora-finetuning}
\resizebox{\columnwidth}{!}{%
\begin{tabular}{lc}
\toprule
\textbf{Hyperparameter} & \textbf{Value} \\
\midrule
\multicolumn{2}{c}{\textit{Model Configuration}} \\
\midrule
Base Models & \begin{tabular}[c]{@{}c@{}}Meta-Llama-3-8B\\Mistral-7B-v0.3\\Gemma-7B\\QWEN-2.5-7B\end{tabular} \\
Load in 8-bit & True \\
\midrule
\multicolumn{2}{c}{\textit{Training Configuration}} \\
\midrule
Sequence Length & 512 \\
Sample Packing & True \\
Pad to Sequence Length & True \\
Validation Set Size & 0.05 \\
Number of Epochs & 10 \\
Gradient Accumulation Steps & 4 \\
Micro Batch Size & 2 \\
Optimizer & adamw\_bnb\_8bit \\
Learning Rate Scheduler & cosine \\
Learning Rate & 2e-4 \\
Weight Decay & 0.0 \\
Warmup Steps & 10 \\
\midrule
\multicolumn{2}{c}{\textit{LoRA Configuration}} \\
\midrule
LoRA Rank & 32 \\
LoRA Alpha & 16 \\
LoRA Dropout & 0.05 \\
Target Linear Layers & True \\
\bottomrule
\end{tabular}
}
\caption{Large \texttt{LM} LoRA Finetuning Hyperparameters}
\label{lora-finetuning}
\end{table}

\begin{table}[h!]
\centering
\resizebox{\columnwidth}{!}{%
\begin{tabular}{lcc}
\toprule
\textbf{Model} & \textbf{Inference Time (s)} \\
\midrule
\multicolumn{2}{c}{\textit{Single Model}} \\
\midrule
OPT-350M & 110.95 \\
OPT-2.7B & 363.71 \\
\midrule
\multicolumn{2}{c}{\textit{HeTLM}} \\
\midrule
OPT-350M ($\alpha$ = 5, $\beta$ = 9, 3 clusters) & 140.55 \\
OPT-350M ($\alpha$ = 3, $\beta$ = 1, 5 clusters) & 163.99 \\
OPT-350M ($\alpha$ = 1, $\beta$ = 0, 6 clusters) & 194.86 \\
\bottomrule
\end{tabular}
}
\caption{Inference Time Comparison on Dataset I}
\label{tab:inference-times}
\end{table}

\subsubsection{Inference Time Comparison}
\label{sec:inference-time}

Table~\ref{tab:inference-times} compares total inference time for different model configurations on Dataset I (10k samples). All measurements were conducted under identical hardware settings with maximum GPU utilization. Notably, our HeTLM approach achieves faster inference times than larger single models while having better performance (Sec.~\ref {results-goog-step1}).

\end{document}